\pdfoutput=1

\documentclass[10pt,twocolumn,letterpaper]{article}

\usepackage{cvpr}
\usepackage{times}
\usepackage{graphicx,epstopdf}
\usepackage{amsmath,eucal,amssymb}
\usepackage{amssymb,bibspacing,url,xspace}

\usepackage[bf,small]{caption}

\usepackage{mdwtab}
\usepackage{multirow}

\usepackage[footnotesize]{subfigure}

\usepackage[]{cite}
\usepackage[]{color}
\usepackage{algorithm2e}
\let\savedalgorithm\algorithm
\let\savedendalgorithm\endalgorithm

\usepackage{algorithm}
\newcommand{\fnorm}[2][2]{\ensuremath{ \left\| #2 \right\|_{ \mathrm{#1} } } }
\newcommand\norm[1]{\left\lVert#1\right\rVert}
\newcommand{\area}{{\rm area}}
\newcommand{\st}{{{\rm s.t.}\!:}\xspace}
\newcommand{\iid}{{i.i.d.}\xspace}
\newcommand{\argmin}{\operatornamewithlimits{argmin}}
\newcommand{\trace}{\operatornamewithlimits{Tr}}
\newcommand{\half}{{\frac{1}{2}}}

\def\bx{{\bf x}}
\def\bw{{\bf w}}
\def\bd{{\bf d}}
\def\ba{{\bf a}}
\def\bb{{\bf b}}
\def\bc{{\bf c}}
\def\bu{{\bf u}}
\def\bmu{{\boldsymbol \mu}}
\def\balpha{{\boldsymbol \alpha}}
\def\bI{{\bf I}}
\def\bS{{\bf S}}
\def\bD{{\bf D}}
\def\bX{{\bf X}}
\def\bA{{\bf A}}
\def\bB{{\bf B}}
\def\bC{{\bf C}}
\def\ones{{\bf 1}}
\def\calR{{\cal R}}

\def\calN{{\cal N}}
\def\st{ {\rm s.t.} }
\def\ADot{ { $\bf \cdot$ } }

\def\T{{\!\top}}

\usepackage[pagebackref=true,breaklinks=true,letterpaper=true,colorlinks,bookmarks=false]{hyperref}
\cvprfinalcopy %

\def\cvprPaperID{} %

\ifcvprfinal\fi
\begin{document}

\title{Online Unsupervised Feature Learning for Visual Tracking}

\author{
Fayao Liu, Chunhua Shen, Ian Reid, Anton van den Hengel\\
School of Computer Science, University of Adelaide, SA 5005, Australia\thanks{Correspondence
should be addressed to C. Shen (email: chunhua.shen@icloud.com).}
}

\maketitle
\thispagestyle{empty}

\begin{abstract}

Feature encoding with respect to an over-complete dictionary learned by unsupervised methods,
followed by spatial pyramid pooling, and linear classification,
has exhibited powerful strength in various vision applications.
Here we propose  to use the feature learning pipeline for visual tracking.
Tracking is implemented using tracking-by-detection and the resulted framework is very simple yet
effective.  First, online dictionary learning is used to build a dictionary, which captures
the appearance changes of the tracking target as well as the background changes.
Given a test image window, we extract local image patches from it and each local patch is encoded
with respect to the dictionary. The encoded features are then pooled over a spatial pyramid to form
an aggregated feature vector. Finally, a simple linear classifier is trained on these features.

Our experiments show that the proposed powerful---albeit simple---tracker, outperforms
all the state-of-the-art tracking methods that we have tested. Moreover, we evaluate the
performance of different dictionary learning and feature encoding methods in the proposed
tracking framework, and analyse the impact of each component in the tracking scenario. We
also demonstrate the flexibility of feature learning by plugging it into Hare et al.'s
tracking method. {\em The outcome is, to our knowledge, the best tracker ever reported}, which
facilitates the advantages of both feature learning and structured output prediction.

\end{abstract}

\section{Introduction}

Robust visual tracking is an important topic in computer vision, with applications
to a wide variety of fields, including video surveillance, motion analysis, object
recognition, \etc. Given the initial state (\eg, bounding box) of a target in a video sequence, a
tracking task aims to infer the states of the target in the succeeding frames.
Despite significant progress made recently
\cite{L1APG,struck,SPOT,CVPR13bLi,CVPR13eYao,tracking-by-detection,ECCV12YAO,Li2011CVPR},
	there still exist challenges from various appearance changes
of the tracking object to diverse background disturbance. The benchmark work of
\cite{benchmark13} identifies
the influential factors of a test sequence to tracking performance into 11 categories,
including illumination variation, occlusion, deformation, motion blur, background
clutters, to name a few.

To address the issue of appearance and background variations, many sophisticated appearance
models have been proposed, which may roughly be categorized into generative and discriminative
based models. Generative models based trackers try to build a robust appearance model of
the tracking object and search for the best matched candidate regions. Examples that fall
into this category are incremental subspace learning \cite{IVT}, sparse representation
based tracking \cite{L1APG,ASLA,SCM,MTT,Li2011CVPR},
distribution fields representation based tracking \cite{DFT},
\etc. In contrast, tracking methods based upon discriminative learning typically model
the tracking object as well as the background, followed by a classification decision to
distinguish the target from its surroundings. Representatives can be the support vector
machines (SVM)  \cite{tracking-by-detection}, boosting ensemble tracking \cite{ensembleTracking}, online
multiple instance learning  \cite{MIT}, bootstrapping binary classifier tracker
\cite{TLD}, structured output tracking \cite{struck}, \etc. These methods usually solve
the tracking problem as detection (tracking-by-detection).
Our proposed tracker applies unsupervised feature
learning in an online fashion to model  both the tracking
target appearance as well as the background, followed by a linear SVM for classification.
Hence it belongs to  this category.

In recent years,  unsupervised feature learning methods have been successfully applied to
many vision tasks such as image classification \cite{TK, ST, beyondPyra}, object recognition
\cite{objectReco}, scene categorization \cite{sceneCategorize}.
 The classical feature learning pipeline mainly consists of three steps: (a) learning
an over-complete dictionary; (b) encoding the  features with the learned dictionary;
(c) spatially pooling the encoded features over a pyramid of regular spatial grids.
The dictionary learning
process is typically unsupervised. Methods such as  K-means,  K-SVD \cite{KSVD},
sparse coding, sparse/denoising autoencoder, or even random sampling,
can be employed.
As for the encoding method, soft threshold, soft assignment, sparse coding, locality-constrained linear
coding \cite{LLC} are commonly applied. It has been shown in \cite{ST} that using
different dictionary learning methods, even random sampling, has little influence on the
classification performance when the dictionary size is sufficiently large,
and the pivotal procedure lies in the encoding step. They
proved that with a simple soft threshold encoding method, state-of-the-art performance can
be achieved in image classification.

The success in those work has inspired us to adapt the image classification pipeline to object tracking.
We highlight the main contributions of this work as follows:
\begin{itemize}
\vspace{-.12cm} \item
We propose a feature learning based tracker using the online dictionary learning method \cite{ODL}.
The online dictionary learning can adapt to  the foreground and background appearance and
effectively update the dictionary words. This is important for online problems like tracking.
Despite the simplicity of the proposed tracker,
it outperforms almost all state-of-the-art trackers in the literature.
\vspace{-.12cm} \item
We evaluate the performance of a few widely-used
dictionary learning and feature encoding methods in the proposed tracking framework.
Due to the nature of tracking problems (such as efficiency requirement and relatively simpler classification
		compared with generic image classification), some helpful conclusions are made, which deviates from the case of
image classification  \cite{ST}.
\vspace{-.12cm} \item
To further demonstrate the superior performance of the learned features over traditional
hand-crafted features in visual tracking, we incorporate the feature learning part into
the Struck tracker \cite{struck} and obtain improved tracking accuracy.
\end{itemize}

\section{Related work}
As a crucial component of the tracking system, the appearance model has been
extensively studied. Besides the traditional hand-crafted features, like texture
\cite{ensembleTracking}, HOG \cite{hogTracking}, Haar-like features \cite{MIT, TLD,
struck}, \etc, the sparse representation has been widely used in tracking, which is
closely related to our feature learning based tracker proposed here.

In \cite{L1Tracker,Li2011CVPR}, the authors
solve the classical sparse coding ($L_1$ minimization) problem to sparsely represent the
tracking object using a set of target templates and trivial templates. Note that their methods, representations
are holistic, and the dictionary is usually constructed using simple methods like sampling or principal component analysis.
In contrast, our method is based on local patches. Also no pooling
is applied in their methods, which can often significantly improve the accuracy, as  shown in our experiments.
In their work, the $L_1$ minimization problem needs to be solved many
times, although \cite{L1APG,Li2011CVPR} applied faster computation to speed up the computation procedure.

 Later, the work of
\cite{SCTracking}  contains learning a dictionary
on SIFT features extracted from general images (\eg, the VOC2010 and Caltech101 datasets) by solving the sparse coding problem,
encoding the feature using the $L_1/L_2$ sparse coding, then applying max-pooling and  training a logistic
regression classifier. In addition to the  aforementioned issue  of extremely expensive computational cost,
		   their method yields
a final representation of high dimension (in their case, it is 14336),
  which can severely hinder its pragmatic
value in tracking. The work of \cite{LSK} proposes to use the histograms of sparse coefficients based on a local
sparse dictionary learned from image patches sampled from the first frame of the sequence
and then applies mean shift for tracking. Similar work can be found in \cite{ASLA, SCM}, although
the work of \cite{ASLA} adopts a different alignment pooling strategy and in \cite{SCM}, it directly concatenates
the learned sparse coefficients instead of pooling. Compared with the methods reviewed above,
we show that using {\em online} dictionary learning with simple but extremely efficient encoding method, rather than
solving the much more expensive $L_1$ minimization problem, we can outperform most state-of-the-art
trackers.

\section{Unsupervised feature learning for tracking}

We follow the well-known tracking-by-detection framework \cite{tracking-by-detection}, which
attempts to learn a classifier to discriminate the target object from its background.
First, we learn a dictionary $\bD=[\bd_1, \bd_2, \ldots, \bd_n] \in \calR^{m \times n}$ of
size $n$ (each column $\bd_j$ denotes a basis\footnote{We call the element in a dictionary basis,
		although it is not necessarily orthogonal.}
vector; if $n>m$, then $\bD$ is over-complete.) based on the image patches\footnote{It
can also be other local descriptors. We simply use
raw pixels of image patches in this work.
We actually found that feature learning on raw pixels usually works better than
feature learning on  low-level image descriptors like local binary patterns.
} extracted from the current frame, and update
it online during the tracking when necessary.

\begin{figure}[t]
\centering
     \includegraphics[width=0.485\textwidth]{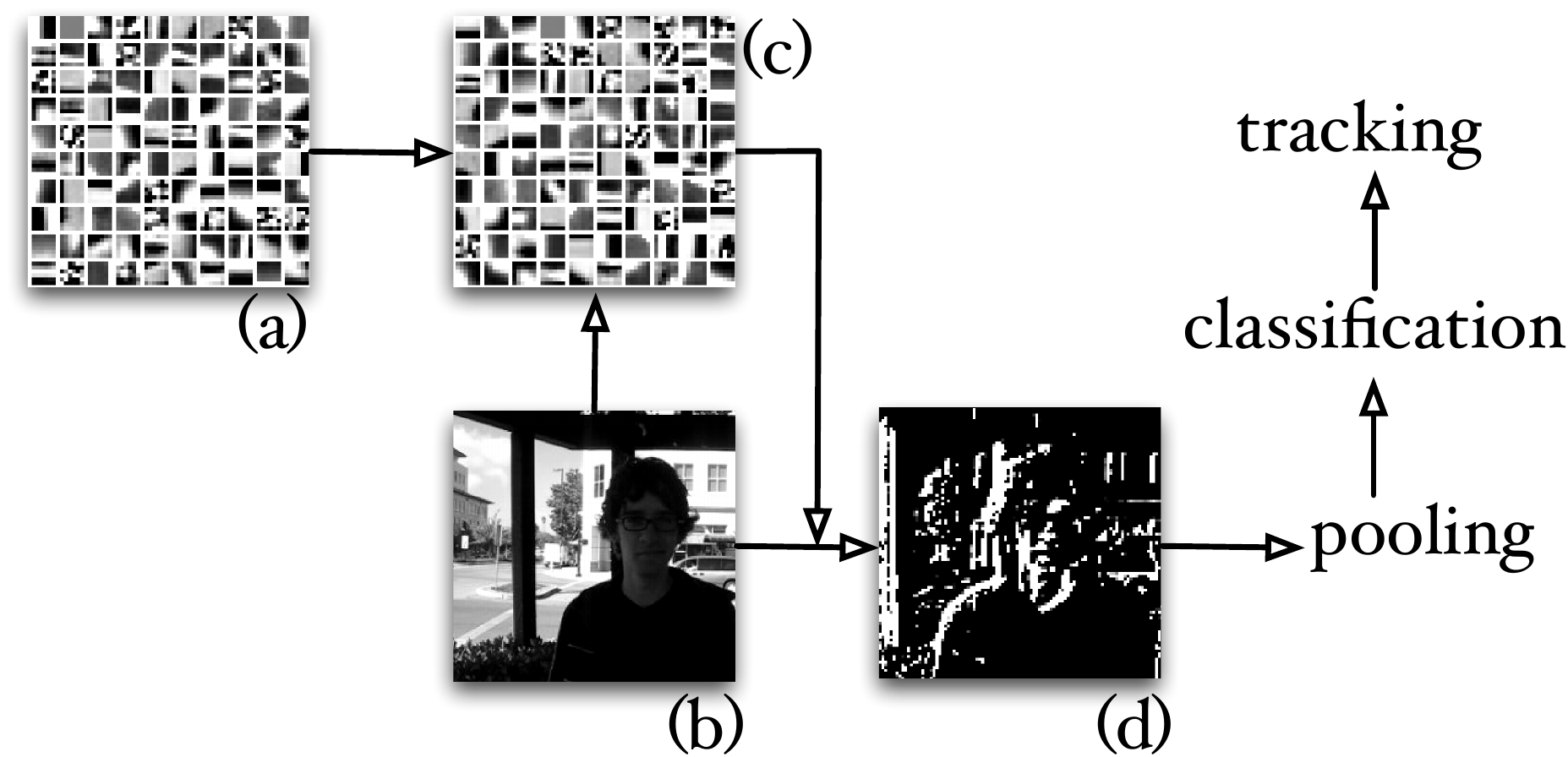}
\caption{An illustration of the pipeline of the proposed feature learning based tracker.
		Component (a) is the current dictionary before update. Before the tracking starts, (a) is learned
		from the  image patches extracted from the first frame.
		(c) is the online updated dictionary based on (a) and the local patches from current frame (b).
		We then encode the local patches in the current frame with respect to the updated dictionary, e.g.,
		using soft threshold encoding. The encoded features (d) are then spatially pooled to form
		the final features, which are used to train a linear classifier. Tracking is implemented using tracking-by-detection.
}  \label{fig:pipeline}
\end{figure}

Due to its efficiency and being easy-to-implement, the soft threshold
(ST) coding strategy is applied here, which writes
$$ \bC = \max\{0, \bD^\T \bX-s\}. $$
Therefore, $ \bC = [\bc_1, \bc_2,
\ldots, \bc_n]^\T \in \calR^{n \times N}
$
are the encoded features, and $s$ is a predefined
threshold. We mainly use soft threshold to encode the original features $\bX=[\bx_1, \bx_2, \ldots, \bx_N] \in
\calR^{m \times N}$ ($\bx_i$ denotes a vector by stacking all pixel values of an image
patch). Then we perform the max-pooling operation to produce the final feature vectors,
which are used to train a linear SVM for detection.
As based on the theoretical and empirical evaluation of \cite{maxPooling},
   max-pooling generally yields more  discriminative features for classification,
   compared to sum or average pooling.
The framework of our feature learning based tracking is illustrated in Figure \ref{fig:pipeline} and the algorithm  is summarized in Algorithm \ref{alg:FLT}.

\subsection{Online dictionary learning}
Various dictionary learning techniques exist in the literature, including K-means, K-SVD
\cite{KSVD}, sparse coding, \etc. Recent studies have shown that using relatively simple
dictionary learning methods, such as K-means or even random sampling, offers surprisingly
promising results in image classification \cite{TK, ST}.
This is true only when the dictionary size is sufficiently large (typically a few thousand),
  which leads to high dimensional feature as the dimension of the feature vector is
linearly proportional to the dictionary size after the encoding process.
For the application of real-time tracking, it requires that the feature dimension cannot be very high
for computational efficiency.
On the other hand, due to temporal changes in the tracking video,
   a fixed dictionary is generally not sufficient to cope with the appearance
changes of the tracking object as well as the background.
We employ  online
dictionary learning of \cite{ODL} to build a relatively small-size dictionary
by taking both the computational efficiency and online update
into consideration.

Given a training set of image patches $\bX$, many classical dictionary learning methods
learn an optimized dictionary $\bD$ by (either exactly or approximately) solving the following objective function:
\begin{equation}
\begin{aligned}
\label{EQ:SC}
\min_{\bD, \balpha} \quad &  \sum_{i=1}^N {\left[
	\frac{1}{2}\fnorm{\bx_i - \bD \balpha_i}^2+\lambda \norm{\balpha_i}_1 \right]},  \\
\st \quad &  \fnorm{\bd_j}^2 \leq 1, \forall j,
\end{aligned}
\end{equation}
where $\balpha=[\balpha_1, \balpha_2, \ldots, \balpha_N] \in \calR^{n \times N}$ are the sparse codes;
$\lambda$ is a regularization parameter;
$\fnorm{\cdot}$ and $\norm{\cdot}_1$ are the $L_2$
and $L_1$ norm respectively. The latter enforces sparsity.
Problem \eqref{EQ:SC} is not jointly convex with respect to $\bD$ and $\balpha$, so it is commonly solved by alternating between the two variables. The online dictionary learning method follows this vein,
		assuming the training set composed of \iid samples.
At each round $t$, the algorithm draws one or more $\bx_t$ ($\bx_{t,1}, \ldots, \bx_{t,\eta}$) and alternates between the classical sparse coding step for computing the sparse code $\balpha_t$ of $\bx_t$ over the dictionary $\bD_{t-1}$,
with the dictionary update step for obtaining $\bD_t$.

The sparse code $\balpha_t$ is solved by the LARS-Lasso \cite{LARS} with $\bD_{t-1}$ fixed:
\begin{align} \label{eq:LARS}
        \balpha_t \triangleq \argmin_{\balpha \in \calR^{n \times \eta}} \frac{1}{\eta} \sum_{j=1}^{\eta} {\left(\frac{1}{2}\fnorm{\bx_{t,j} - \bD_{t-1} \balpha_j}^2+\lambda \norm{\balpha_j}_1 \right)}.
        \end{align}

While the dictionary is updated by optimizing:
\begin{equation}
\begin{aligned} \label{eq:odl}
\bD_t &\triangleq \argmin_{\fnorm{\bD}^2 \leq 1} \frac{1}{t\eta}\sum_{i=1}^t\sum_{j=1}^{\eta} \left( \half \fnorm{\bx_{i,j} - \bD \balpha_{i,j}}^2 + \lambda \norm{\balpha_{i,j}}_1 \right), \\
&= \argmin_{\fnorm{\bD}^2 \leq 1} \frac{1}{t} \left( \half \trace(\bD^\T \bD \bA_t) - \trace(\bD^\T \bB_t) \right),
\end{aligned}
\end{equation}
with
\begin{equation}
\begin{aligned} \label{eq:AB}
 \bA_t=&\frac{1}{\eta}\sum_{i=1}^t\sum_{j=1}^{\eta} \balpha_{i,j} \balpha_{i,j}^\T \triangleq [\ba_1^{(t)}, \ldots, \ba_n^{(t)}], \\
 \bB_t=&\frac{1}{\eta}\sum_{i=1}^t\sum_{j=1}^{\eta} \bx_{i,j} \balpha_{i,j}^\T \triangleq [\bb_1^{(t)}, \ldots,
\bb_n^{(t)}],
\end{aligned}
\end{equation}
both of which are also updated online.
Here $ \trace (\cdot )$ denotes the trace of a matrix, and $\bA_t \in \calR^{n \times n}$, $\bB_t \in \calR^{m \times n}$.
The optimization problem \eqref{eq:odl} is solved by sequentially updating the $j$-th column of $\bD$ through an orthogonal projection onto the constrained set:
 \begin{equation}
        \begin{aligned} \label{eq:UpdateDict}
        \bd_j^{(t)} \leftarrow \frac{1}{\max(\fnorm{\bu_j}, 1)} \bu_j,
        \end{aligned}
        \end{equation}
where $\bu_j = \frac{1}{\bA_t[j,j]} \left( \bb_j^{(t)} - \bD_{t-1} \ba_j^{(t)} \right) + \bd_j^{(t-1)}$, with $\bA_t[j,j]$ denotes the $j$-th row and $j$-th column element of $\bA_t$, and $\bd_j^{(t-1)}$, $\bd_j^{(t)}$ is the $j$-th column of $\bD_{t-1}$ and $\bD_t$ respectively.

The algorithm
is summarized in Algorithm \ref{alg:ODL}. It is worth noting that the method can also be
used in an off-line fashion to train on fixed-size data by cycling over a randomly permuted
training set to draw $\bx_t$. In the tracking task, the dictionary can be off-line learned
from natural images or the first frame of the sequence. We provide a comparison
of these three cases in the experiment section.

\begin{figure*}[t]
\centering
     \includegraphics[width=0.8\textwidth, height=0.23\textwidth]{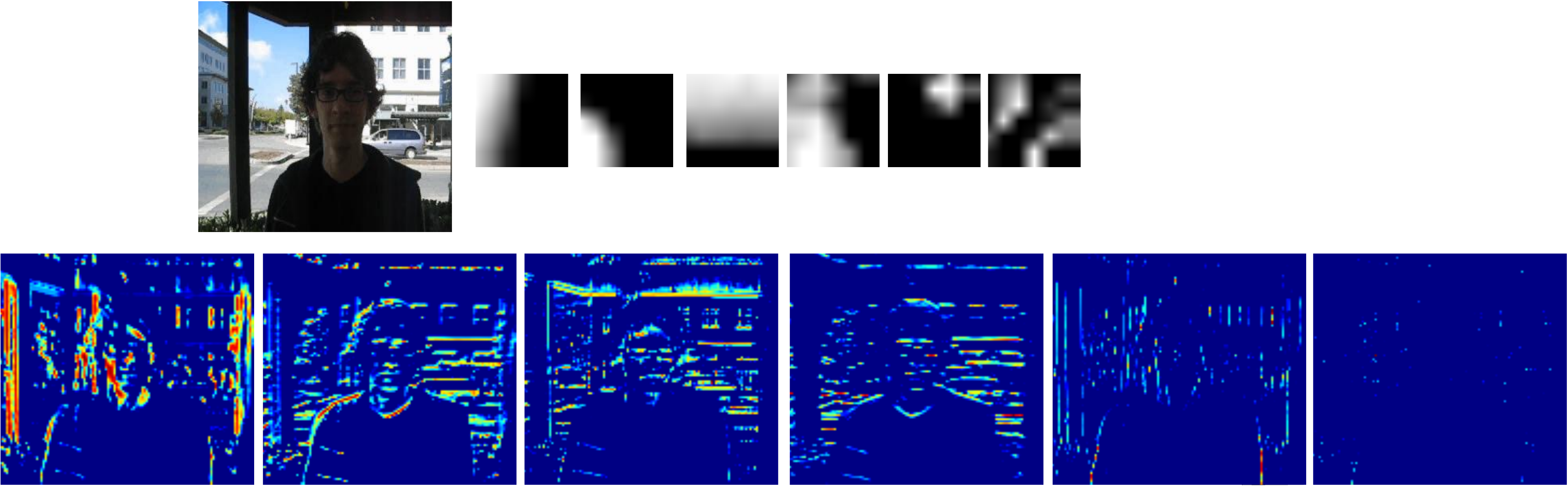}
\caption{The image frame and the learned bases (first row) ordered by the weighting skeme from most important to least important and their corresponding encoded features/responses (second row). The rank is 1st, 20th, 40th, 60th, 80th, 100th from left to right.}  \label{fig:basisVSresp}
\end{figure*}

\paragraph{Dictionary update}
To avoid the unstable performance caused by too frequent update as well as to ensure
efficiency, we apply some heuristic strategies here.
To capture the appearance change of the object, we introduce a weighting scheme for each basis in $\bD$, which is defined as the normalized $L_2$ norm of the encoded features. Specifically, the $j$-th basis $\bd_j$ is weighted as $\frac{\fnorm{\bc_j}}{\sum_{j=1}^n{\fnorm{\bc_j}}}$. It indicates the relative importance of the bases in the encoding process, and essentially the appearance of the region. According to this weighting scheme, we can sort the bases from the most important to the least important. During the tracking, if the overlap between the top half bases of the two detected target regions in consecutive frames below a threshold ($0.9$ in our experiment), then there is possibly appearance change happening, and the dictionary is updated. We give an illustration by visualizing the ordered learned bases (100 in total) with their corresponding encoded responses in figure \ref{fig:basisVSresp}. As can be seen, the ranked bases provide some intuitive insights into the feature learning approach.

\setcounter{AlgoLine}{0}
\linesnumbered\SetVline
\begin{algorithm}[t]
\caption{Online feature learning based tracking.} \label{alg:FLT}
\centering
{%
   \begin{minipage}[]{0.94\linewidth}
    \KwIn{
    Initial dictionary $\bD_0$; image patch size $p$; step size $q$; length of sequence $T$.
    }
	{ {\bf Initialize}:
        $\bA_0 \leftarrow 0, \bB_0 \leftarrow 0$.
   }

   {\bf for} $t=1$ to $T$ {\bf do}
   {

    \ADot
        Extract image patches $\bx_{t,1}, \ldots, \bx_{t,\eta}$ of size $p \times p$ at step size $q$ from frame $t$ and do contrast normalization.

    \ADot
        Update dictionary $\bD_t$ by calling Algorithm \ref{alg:ODL}.

    \ADot
    		Sample a bunch of boxes around the previous estimation of the tracking target.

	\ADot
		Encode the raw pixel features of the patches extracted within each sampled boxes using $\bD_t$ by soft threshold coding.

	\ADot
		Perform max-pooling over a spatial pyramid of multiple layers.

    \ADot
    		Train an LS-SVM by solving \eqref{eq:lssvm_solution}. %

    	\ADot
    		Predict the most confident bounding box of the tracking target.
   }

   {\bf end for}

\end{minipage}
}
\end{algorithm}

\setcounter{AlgoLine}{0}
\linesnumbered\SetVline
\begin{algorithm}[t]
\caption{Dictionary update.} \label{alg:ODL}
\centering
{%
   \begin{minipage}[]{0.94\linewidth}
    \KwIn{
    Training samples $\bx_{t,1}, \ldots, \bx_{t, \eta} \in \calR^m$; regularization parameter $\lambda$; $\bA_{t-1}$; $\bB_{t-1}$; $\bD_{t-1}$.
    }

	\ADot
        Solve Eq. \eqref{eq:LARS} for $\balpha$.

    \ADot
        Update $\bA_t$: $\bA_t \leftarrow  \bA_{t-1} + \frac{1}{\eta}\sum_{i=1}^{\eta}\balpha_{t, i} \balpha_{t,i}^\T$;  %

	\ADot
		Update $\bB_t$: $\bB_t \leftarrow \bB_{t-1} + \frac{1}{\eta}\sum_{i=1}^{\eta}\bx_{t, i} \balpha_{t,i}^\T$;

   \ADot
   {\bf for} $j=1$ to $n$ {\bf do}
	{

    \ADot
      \quad Update the $j$-th column of $\bD_{t-1}$ by \eqref{eq:UpdateDict};

	\ADot
      {\bf end for}
    };

\KwOut{
    The updated dictionary $\bD_t$. %
}

\end{minipage}
}
\end{algorithm}

\subsection{Re-training of linear classification}

To build an appearance discriminative model, we train a linear least-squared SVM (LS-SVM)
classifier on the learned features, mainly due to its fast closed-form solution.
Of course many other classifiers can be used here.

Given a set of training examples $\{\bx_i,
y_i \}_{i=1}^N $, where $\bx_i \in \calR^m$ and $y_i \in \{-1, +1\}$, the LS-SVM
learns a classifier $f(\bx)=\bw^\T\bx+b$ by optimizing the following objective function
\cite{lssvm}:
\begin{align}
\label{EQ:lssvm}
            \min_{ \bw, b }\sum_{i=1}^N \fnorm{f(\bx_i)-y_i}^2 + \gamma \fnorm{\bw}^2,
\end{align}
where $\fnorm{\cdot}$ is the $L_2$ norm and $\gamma$ is the trade-off parameter. To simplify notation, we define $\ones$ as an vector of all ones, $\bX=[\bx_1, \bx_2, \ldots, \bx_N]$ to be the data matrix, $N_{+}, N_{-}$ be the positive and negative sample number respectively, $\bmu_{+}, \bmu_{-}$ be the positive and negative sample mean, and $\bmu$ be the mean of all training samples. Obviously we have $N=N_{+}+N_{-}$ and $\bmu=\frac{N_{+}}{N}\bmu_{+}+\frac{N_{-}}{N}\bmu_{-}$.  Then the closed form solution of \eqref{EQ:lssvm} can be formulated as:
\begin{equation}
\begin{aligned} \label{eq:lssvm_solution}
\bw &= \frac{2N_{+}N_{-}}{N^2}(\bS+\frac{\gamma}{N}\bI)^{-1}(\bmu_{+}-\bmu_{-}), \\
b&=\frac{N_{+}N_{-}}{N}-\bmu^\T\bw,
\end{aligned}
\end{equation}
where $\bI$ is an identity matrix and $\bS$ is the covariance matrix formulated as
$\bS=\frac{1}{N}(\bX-\bmu \ones^\T)(\bX-\bmu \ones^\T)^\T$. During tracking, we use
an online reservoir of boxes from a maximum number of frames (30 in our experiment)
for training. Generally, the earliest tracking results are more accurate, while the
latest ones capture the recent appearance of the tracking target. Based on these two
considerations, we select the boxes from the first 10 together with the most recent 20
frames to maintain the reservoir.

\section{Experiments}

In this section, we offer a comprehensive evaluation of the proposed tracker on
twenty sequences, most of which can be found at the website of the first author of
\cite{benchmark13}. These sequences contain various challenging situations in object
tracking, like illumination variation, occlusion, deformation, background clutters, fast
motion \etc. For a detailed attribute description, please refer to \cite{benchmark13}.
Two widely-used evaluation criteria are utilized here, namely, the center location
error (CLE) and the PASCAL VOC overlap ratio (VOR), with the latter defined as ${\rm
VOR}={\area(S_T \bigcap S_{GT})}/{\area(S_T \bigcup S_{GT})}$, where $S_T$ is the
tracking result box and $S_{GT}$ the ground truth bounding box.

We use a search radius of 30 for tracking and 60 for training classifier, as did in Struck
\cite{struck}. The dictionary is initially learned from image patches of the first frame
and then online updated. We extract $8\times8$ patches at a step size 4 for large tracking
objects and $6\times6$ with stride 2 for small targets. The patches are then normalized by
subtracting the mean and dividing by the standard deviation for contrast normalization.
Note that we do not do the unit length normalization here as it degrades performance. We
use a dictionary size of 100 ($n=100$) and soft threshold (ST) coding with three-level
max pooling ($1\times1, 2\times2, 3\times3$), which yields a feature dimension of 1400.
As we do not do the unit length normalization, we empirically set the threshold of ST as
$s=0.25 \cdot \max{(\bD^\T \bX)}$ and use it throughout all the sequences.
We use the  optimization toolbox \cite{ODL}
for online updating the dictionary and solving the sparse coding problem.

During the tracking, we maintain a reservoir of 30 frames (the first 10 and the most
recent 20; fixed for all the sequences) for re-training the LS-SVM.
The classifier is initially trained with the first two labelled frames and updated every four frames.
Our unoptimized Matlab implementation runs around 4 frames per second with no dictionary
updating and around 2.5 frames per second with dictionary update,
		 on a standard PC machine using a single core.

\subsection{Comparison with state-of-the-art trackers}
We first compare our tracker with eight state-of-the-art trackers, which are Struck
(structured output tracker \cite{struck}), SCM (sparsity-based collaborative model
\cite{SCM}), ASLA (adaptive structural local appearance model \cite{ASLA}), L1APG ($L_1$
tracker using accelerated proximal gradient approach \cite{L1APG}), DFT (distribution
field tracker \cite{DFT}), MTT (multi-task sparse learning tracker \cite{MTT}), TLD
(bootstrapping binary classifier tracker \cite{TLD}), IVT (incremental subspace tracker
\cite{IVT}).
The publicly available benchmark code of \cite{benchmark13} with initial settings are used
for evaluating their results.
We report the average VORs and CLEs in Table \ref{tab:overlapScore} and Table
\ref{tab:CLE} respectively. For our method, due to the randomness introduced by the
dictionary learning process, we run 5 times and report the median results. The results of
our tracker both with and without dictionary update process are included in the table.
From the results, we can see that our tracker with online dictionary update achieves
the best overall performance across all the twenty sequences, especially on the david3,
box, iceball and bolt, where the other trackers lose the target at different frames. One
more notable conclusion is that even without dictionary update, our tracker performs
surprisingly well, which may result from the fact that most tracking scenes consist of
relatively simple image patterns. We will give more discussions on the dictionary update
later.

\begin{table*}
\centering
\resizebox{.96\linewidth}{!}
{
\begin{tabular}{  r | c | c | c | c | c | c | c | c | c | c }
\hline
	{{Sequence}} & {Ours} & {Ours\_U} & {Struck} \cite{struck} & {SCM} \cite{SCM} & {ASLA} \cite{ASLA} & {L1APG} \cite{L1APG}
& {DFT} \cite{DFT} & {MTT} \cite{MTT} & {TLD} \cite{TLD} & {IVT} \cite{IVT} \\ \hline
	\hline
	\emph{david} & \textcolor{blue}{0.85} & \textcolor{red}{0.86} & 0.79 & 0.61 & 0.59 & 0.40 & 0.48 & 0.30 & 0.59 & 0.47 \\ %
	\emph{girl} & 0.79 & \textcolor{red}{0.81} & \textcolor{blue}{0.80} & 0.42 & 0.63 & 0.73 & 0.39 & 0.62 & 0.42 & 0.04 \\ %
	\emph{faceocc1} & 0.80 & 0.80 & 0.83 & \textcolor{red}{0.92} & \textcolor{blue}{0.86} & 0.78 & 0.51 & 0.69 & 0.65 & 0.78 \\ %
	\emph{faceocc2} & \textcolor{blue}{0.81} & \textcolor{red}{0.82} & 0.74 & 0.74 & 0.74 & 0.75 & \textcolor{red}{0.82} & 0.75 & 0.48 & 0.66 \\ %
	\emph{david3} & \textcolor{red}{0.73} & \textcolor{red}{0.73} & 0.29 & 0.48 & 0.49 & 0.38 & \textcolor{blue}{0.56} & 0.10 & 0.28 & 0.49 \\ %
	\emph{woman} & 0.71 & 0.72 & \textcolor{blue}{0.74} & 0.32 & 0.15 & 0.16 & \textcolor{red}{0.76} & 0.17 & 0.28 & 0.14 \\ %
	\emph{shaking} & \textcolor{red}{0.71} & \textcolor{red}{0.71} & 0.08 & 0.55 & \textcolor{blue}{0.64} & 0.36 & 0.14 & 0.55 & 0.12 & 0.03 \\ %
	\emph{fskater} & \textcolor{blue}{0.80} & \textcolor{red}{0.81} & \textcolor{red}{0.81} & 0.59 & 0.73 & 0.67 & 0.67 & 0.77 & 0.51 & 0.62 \\ %
	\emph{bird2} & \textcolor{red}{0.78} & \textcolor{red}{0.78} & 0.55 & \textcolor{blue}{0.75} & 0.51 & 0.43 & 0.74 & 0.08 & 0.23 & 0.48 \\ %
	\emph{deer} & 0.73 & \textcolor{red}{0.75} & \textcolor{blue}{0.74} & 0.10 & 0.05 & 0.60 & 0.26 & 0.66 & 0.25 & 0.03 \\ %
	\emph{dollar} & 0.82 & 0.82 & 0.70 &0.85 & \textcolor{red}{0.87} & 0.80 & \textcolor{blue}{0.86} & \textcolor{red}{0.87} & 0.30 & \textcolor{red}{0.87} \\ %
	\emph{box} & \textcolor{blue}{0.81} & \textcolor{red}{0.82} & 0.34 & 0.17 & 0.13 & 0.22 & 0.34 & 0.26 & 0.52 & 0.55 \\ %
	\emph{board} & \textcolor{blue}{0.81} & \textcolor{red}{0.82} & 0.76 & 0.36 & 0.69 & 0.08 & 0.34 & 0.19 & 0.24 & 0.15 \\ %
	\emph{coke11$^*$} & \textcolor{blue}{0.55} & \textcolor{red}{0.59} & \textcolor{blue}{0.55} & 0.50 & 0.18 & 0.10 & 0.16 & 0.43 & 0.41 & 0.05 \\ %
	\emph{tiger1$^*$} & \textcolor{red}{0.74} & \textcolor{blue}{0.73} & \textcolor{blue}{0.71} & 0.12 & 0.29 & 0.44 & 0.36 & 0.38 & 0.39 & 0.09 \\ %
	\emph{tiger2$^*$} & \textcolor{red}{0.74} & \textcolor{blue}{0.73} & 0.59 & 0.27 & 0.13 & 0.26 & 0.57 & 0.30 & 0.31 & 0.13 \\ %
	\emph{sylvester$^*$} & \textcolor{red}{0.75} & \textcolor{blue}{0.73} & 0.69 & 0.56 & 0.59 & 0.41 & 0.50 & \textcolor{blue}{0.71} & 0.66 & 0.54 \\ %
	\emph{trellis$^*$} & \textcolor{blue}{0.78} & 0.77 & \textcolor{red}{0.79} & 0.61 & 0.75 & 0.43 & 0.37 & 0.44 & 0.33 & 0.15 \\ %
	\emph{iceball$^*$} & \textcolor{blue}{0.70} & \textcolor{red}{0.72} & 0.51 & 0.48 & 0.45 & 0.49 & 0.08 & 0.06 & 0.07 & 0.06 \\ %
	\emph{bolt$^*$} & \textcolor{blue}{0.71} & \textcolor{red}{0.73} & 0.01 & 0.02 & 0.01 & 0.01 & 0.03 & 0.01 & 0.16 & 0.01 \\ \hline
	\emph{average} &\textcolor{blue}{0.75}	 &\textcolor{red}{0.76}	&0.60 &0.47	&0.47	&0.43	&0.45	&0.42 	&0.36	 &0.32 \\
	\hline

\end{tabular}
}
\caption{Compared average VORs on 20 sequences. Ours\_U and Ours refer to our tracker with and without dictionary update respectively. The last row are the averaged results on all sequences. The best and second best are shown in red and blue respectively. The sequences marked with star (*) are evaluated by using a patch size of 6 and stride 2 due to small target objects and all the other sequences use a patch size of 8 with stride 4.} \label{tab:overlapScore}
\end{table*}

\begin{table*}
\small
\center
\resizebox{.96\linewidth}{!}
{
\begin{tabular}{  r | c | c | c | c | c | c | c | c | c | c  }
\hline
	{{Sequence}} & {Ours} & {Ours\_U} & {Struck} \cite{struck} & {SCM} \cite{SCM} & {ASLA} \cite{ASLA} &
	{L1APG} \cite{L1APG} & {DFT} \cite{DFT} & {MTT} \cite{MTT} & {TLD} \cite{TLD} & {IVT} \cite{IVT} \\ \hline
	 \hline
	\emph{david} & 5.7 & 5.2 & 8.8 & 5.8 & \textcolor{red}{2.9} & 44.9 & 55.8 & 93.8 & 8.1 & \textcolor{blue}{3.9} \\ %
	\emph{girl} & 12.1 & \textcolor{blue}{10.4} & \textcolor{red}{10.2} & 60.5 & 30.3 & 13.1 & 51.2 & 23.3 & 31.7 & 145.5 \\ %
	\emph{faceocc1} & 12.0 & 11.6 & 7.5 & \textcolor{red}{4.1} & \textcolor{blue}{7.1} & 13.0 & 47.3 & 20.6 & 19.0 & 12.8 \\ %
	\emph{faceocc2} & 9.1 & 8.7 & 9.2 & 8.4 & 9.3 & \textcolor{blue}{8.2} & 8.5 & 10.1 & 16.3 & \textcolor{red}{7.9} \\ %
	\emph{david3} & \textcolor{blue}{10.3} & \textcolor{red}{10.1} & 106.7 & 75.6 & 85.6 & 90.0 & 51.0 & 399.2 & 135.7 & 51.6 \\ %
	\emph{woman} & 5.4 & \textcolor{blue}{5.2} & \textcolor{red}{4.1} & 118.8 & 157.7 & 128.7 & 8.5 & 138.1 & 78.9 & 181.6 \\ %
	\emph{shaking} & \textcolor{red}{10.6} & \textcolor{blue}{10.8} & 123.9 & 18.1 & 11.6 & 84.5 & 174.8 & 18.2 & 65.6 & 86.7 \\ %
	\emph{fskater} & 10.5 & \textcolor{blue}{9.2} & \textcolor{red}{7.1} & 26.3 & 8.1 & 21.2 & 22.8 & 12.3 & 15.9 & 19.4 \\ %
	\emph{bird2} & \textcolor{red}{7.6} & \textcolor{blue}{7.7} & 20.7 & 8.7 & 22.1 & 57.6 & 10.7 & 145.6 & 75.1 & 30.7 \\ %
	\emph{deer} & 8.2 & \textcolor{blue}{7.9} & \textcolor{red}{5.3} & 108.3 & 144.2 & 24.2 & 98.7 & 11.9 & 117.7 & 179.4 \\ %
	\emph{dollar} & 7.5 & 6.6 & 14.7 & 5.2 & \textcolor{red}{4.2} & 6.9 & \textcolor{blue}{5.0} & 5.2 & 70.0 & \textcolor{blue}{5.0} \\ %
	\emph{box} & 9.4 & 8.9 & 140.0 & 127.1 & 165.7 & 104.1 & 106.2 & 100.6 & 20.7 & 18.3 \\ %
	\emph{board} & \textcolor{blue}{16.7} & \textcolor{red}{15.5} & 24.0 & 100.9 & 34.3 & 220.0 & 98.3 & 142.8 & 130.8 & 157.4 \\ %
	\emph{coke11$^*$} & \textcolor{blue}{8.1} & \textcolor{red}{7.4} & 8.3 & 10.9 & 29.5 & 64.1 & 30.2 & 17.9 & 14.1 & 44.5 \\ %
	\emph{tiger1$^*$} & \textcolor{red}{5.4} & \textcolor{blue}{5.7} & 6.0 & 86.1 & 32.9 & 23.2 & 30.5 & 26.9 & 22.4 & 60.3 \\ %
	\emph{tiger2$^*$} & \textcolor{blue}{6.4} & \textcolor{red}{6.2} & 9.2 & 25.9 & 41.2 & 35.4 & 12.5 & 24.3 & 17.7 & 44.7 \\ %
	\emph{sylvester$^*$} & \textcolor{red}{6.7} & \textcolor{blue}{7.1} & 8.4 & 19.8 & 21.0 & 41.9 & 36.1 & \textcolor{blue}{7.1} & 8.6 & 36.7 \\ %
	\emph{trellis$^*$} & \textcolor{blue}{5.2} & 6.2 & \textcolor{red}{5.0} & 15.0 & 7.1 & 41.6 & 54.6 & 43.9 & 41.0 & 156.7 \\ %
	\emph{iceball$^*$} & \textcolor{blue}{4.9} & \textcolor{red}{4.5} & 15.6 & 32.0 & 18.3 & 14.4 & 116.6 & 137.3 & 101.5 & 106.0 \\ %
	\emph{bolt$^*$} & \textcolor{blue}{6.9} & \textcolor{red}{6.6} & 365.2 & 374.6 & 385.3 & 408.4 & 367.3 & 485.6 & 88.0 & 379.4 	 \\ \hline
	\emph{average} &\textcolor{blue}{8.4}	&\textcolor{red}{8.1} &45.0	&61.6	&60.9	&72.3	&69.3	&93.2	&53.9	&86.4 \\
	\hline

\end{tabular}
}
\caption{Compared average CLEs in pixels on 20 sequences. Ours\_U and Ours refer to our tracker with and without dictionary update respectively. The last row are the averaged results on all sequences. The best and second best are shown in red and blue respectively. The sequences marked with star (*) are evaluated by using a patch size of 6 and stride 2 due to small target objects and all the other sequences use a patch size of 8 with stride 4.} \label{tab:CLE}
\end{table*}

\subsection{Analysis of   feature learning}
In this section, we examine several factors that have impact on the performance of the proposed tracker.

\paragraph{Evaluation of different dictionary learning methods}
We compare the online dictionary learning algorithm \cite{ODL} used in this paper with
two other typical dictionary learning methods, namely, K-means and K-SVD \cite{KSVD}.
We also include results using random sampled (RS) patches as dictionary, and all the
methods use the image patches extracted from the first frame. One may suspect using
patches obtained from natural images may yield better performance, as they may provide
more general patterns. To justify this point, we also run the ODL method by utilizing
100000 image patches randomly selected from a segmentation database and use it through
out all the sequences. The dictionary size is fixed at 100 for all the methods. Table
\ref{tab:DLComp} shows the average VORs and CLEs on eight sequences. The results indicate
that the random sample method performs bad in the case of small dictionary size, and using
different dictionary learning methods has little influence in the tracking performance,
which is in accordance with the conclusion of \cite{ST} in image classification. The
reason why we use ODL rather than the other two is that K-SVD is more time consuming and
K-means suffers from unstable performance in case of online update. One more conclusion
can be made from Table \ref{tab:DLComp} is that using image patches directly from the
sequence can better capture the patterns of the tracking object as well as the background,
especially when the dictionary size is not large enough.

\begin{table*} \center
\resizebox{.65\linewidth}{!}
{
\begin{tabular}{ r | c  c | c  c | c  c | c  c | c  c  }
\hline
\multirow{2}{*}{{{Sequence}}} &\multicolumn{2}{c|}{\bf{ODL\_G}} &\multicolumn{2}{c|}{\bf{ODL}} &\multicolumn{2}{c|}{\bf{K-means}} &\multicolumn{2}{c|}{\bf{K-SVD}} &\multicolumn{2}{c}{\bf{RS}}	\\
&{VOR} &{CLE} &{VOR} &{CLE} &{VOR} &{CLE} &{VOR} & {CLE} & {VOR} & {CLE} \\
\hline
\emph{bird2}	&0.72 &10.8 &\textbf{0.78} &\textbf{7.6} 	&0.74 &9.5	&0.73 &10.0	 &0.71 &11.4\\
\emph{tiger2}	&0.72 &\textbf{6.2}	&\textbf{0.74} &6.4	&0.72 &6.5	&0.71 &6.3 &0.65 &11.9\\
\emph{david}	&0.83 &6.6	&0.85 &5.7	&0.85 &5.6	&\textbf{0.86} &\textbf{5.3}	 &0.79 &8.0\\
\emph{fskater}	&0.75 &14.1	&0.80 &10.5 	&\textbf{0.81} &\textbf{9.4}	&0.78 &10.6	 &0.70 &15.5\\
\emph{faceocc2} 	&\textbf{0.82} &\textbf{8.7} &8.7 &9.1	&0.80 &10.3	&0.79 &10.9	 &0.81 &9.0\\
\emph{dollar}	&0.78 &9.2	&0.82 &7.5	&0.79 &9.3	&\textbf{0.83} &\textbf{6.7}	&0.67  &14.6\\
\emph{board}	&0.78 &19.5 	&0.81 &16.7 	&\textbf{0.82} &\textbf{16.0} 	&0.80 &17.8	 &0.47 &106.8\\
\emph{trellis}	&0.77 &5.8	&\textbf{0.78} &\textbf{5.2} &0.78 &5.6	&0.78 &5.4	&0.69 &12.8\\
\hline
\emph{average} &0.77 &10.1 &\bf{0.80} &\textbf{8.5} &0.79 &9.0 &0.79 &9.1  &0.69 &23.8\\
\hline
\end{tabular}
}
\caption{Performance comparison
	(VORs and CLEs) of different dictionary learning methods. ODL\_G and ODL refer to online dictionary learning with general patches extracted from natural images and the first frame of the sequence respectively. RS denotes random sample. K-means, K-SVD and RS all use image patches from the first frame of the sequence.
		The last row shows the results averaged over all sequences.
		} \label{tab:DLComp}
\end{table*}

\paragraph{Evaluation of different encoding schemes}
Besides the soft threshold (ST) and sparse coding (SC), there are several encoding schemes
exist in the literature, which include soft assignment (SA), localized soft assignment
(LSA) \cite{LSA} and triangle K-means (TK) \cite{TK} \etc. Given a learned dictionary
$\bD$, the encoding process provides a feature mapping from $\bX$ to $\bC$. We summarize
the formulations of the five encoding methods in Table \ref{tab:encoding}.
After obtaining the dictionary using online dictionary learning, we compare the tracking
results with the five different encoding methods. The threshold $t$ in ST is empirically
chose as $0.25 \cdot \max{(\bD^\T \bX)}$.
The smoothing factor $\beta$ in SA and LSA is set to 10 as suggested in \cite{LSA} and
the neighborhood size $k$ in LSA is tuned from $\{5, 10, 20\}$. The trade-off parameter
$\lambda$ in SC is optimally chose from $\{0.25, 0.5, 1, 1.5\}$.
Table \ref{tab:codingComp} reports the average VORs and CLEs on eight sequences. As can
be observed, the simple soft threshold encoding performs on par with sparse coding, while
better than the other three. While sparse coding needs to solve an $L_1$-regularized
linear least-squares problem every time (with fixed dictionary), soft threshold coding
only requires a max operation.

\begin{table*}
\small
\center
\resizebox{.96\linewidth}{!}
{
\begin{tabular}{ l  c  c  c  c c }
\hline
& {ST} & {TK} & {SA} & {LSA} & {SC} \\
\hline
 \multirow{2}{*}{$c_j$} & \multirow{2}{*}{$\max\{0, \bd_j^\T \bx -s \}$}
&\multirow{2}{*}{$\max\{0, \bmu - \fnorm{\bx - \bd_j} \}$}
&\multirow{2}{*}{$\frac{\exp(-\beta\fnorm{\bx-\bd_j}^2)}{\sum_{l=1}^n\exp{(-\beta\fnorm{\bx-\bd_l}^2)}}$}
&$\frac{\exp(-\beta\fnorm{\bx-\bd_j}^2)}{\sum_{l=1}^n\exp{(-\beta\fnorm{\bx-\bd_l}^2)}}$ if $\bd_j \in \calN_k(\bx)$
&\multirow{2}{*}{$\max\{ 0, \balpha \}$} \\
& & & &$0$ otherwise  & \\
\hline
\end{tabular}
}
\caption{Formulations of five different encoding schemes. $c_j$ is the $j$-th encoded features of $\bx$ with respect to the $j$-th basis $\bd_j$; $\bmu$ is the mean of $\fnorm{\bx - \bd_j}$ over all $j$; $\beta$ is the smoothing factor; $\calN_k(\bx)$ represents the $k$-nearest neighborhood of $\bx$ defined by the distance $\fnorm{\bx-\bd_j})$; $\balpha$ is learned by solving \eqref{EQ:SC} for $\bx$ with $\bD$ fixed. }
\label{tab:encoding}
\end{table*}

\begin{table*} \center
\resizebox{.65\linewidth}{!}
{
\begin{tabular}{ r | c  c | c  c | c  c | c  c | c  c  }
\hline
\multirow{2}{*}{{{Sequence}}} &\multicolumn{2}{c|}{{ST}} &\multicolumn{2}{c|}{{TK}} &\multicolumn{2}{c|}{{SA}}
&\multicolumn{2}{c|}{{LSA}} &\multicolumn{2}{c}{{SC}}	\\
&{VOR} &{CLE} &{VOR} &{CLE} &{VOR} &{CLE} &{VOR} &{CLE} &{VOR} &{CLE} \\
\hline
	\emph{bird2}  &\textbf{0.78} &\textbf{7.6}	&0.74 &9.3	&0.77 &8.1	&0.75 &8.8	&0.75 &9.8 \\
	\emph{tiger2} 	&0.74 &6.4	&0.70 &5.7	&0.73  &6.4 	&\textbf{0.76} &\textbf{5.3}	&0.75 & 5.9\\
	\emph{david}	&0.85 &5.7	&0.85 &5.9	&0.83 &6.6	&0.85 &5.9	&\textbf{0.86} &\textbf{5.3}\\
	\emph{fskater}	&0.80 &10.5 	&0.80 &\textbf{8.4}	&0.77 &10.4 	&0.78 &10.7 	&\textbf{0.81} &9.4\\
	\emph{faceocc2}	&0.81 &9.1	&0.81 &9.2	&0.75 &12.2 	&0.79 &10.2	&\textbf{0.82} &\textbf{8.4}\\
	\emph{dollar}	&0.82 &7.5	&0.81 &8.6	&\textbf{0.83} &\textbf{6.4} 	&0.82 &7.0	 &0.81 &7.2\\
	\emph{board}	&0.81 &16.7 	&\textbf{0.81} &\textbf{15.8}	&0.76 &22.5 	&0.80 &17.4	&0.79 &18.4	\\
	\emph{trellis}	&\textbf{0.78} &\textbf{5.2} 	&0.72 &8.0	&0.76 &6.2	&0.76 &6.3	 &0.77 &6.4 \\ \hline
	\emph{average} &\bf{0.80} &\textbf{8.5} &0.78 &8.9 &0.78 &9.9 &0.79 &9.0 &\bf{0.80} &8.9\\
	\hline
\end{tabular}
}
\caption{Performance comparison of different encoding schemes (VORs and CLEs).
	ST: soft threshold; TK: triangle K-means; SA: soft assignment; LSA: localized soft assignmetn; SC: sparse coding.
		The last row shows the averaged results over all sequences.
		The best results are showed in bold.} \label{tab:codingComp}
\end{table*}

\begin{figure}[t]
\centering
     \includegraphics[width=0.4985\textwidth]{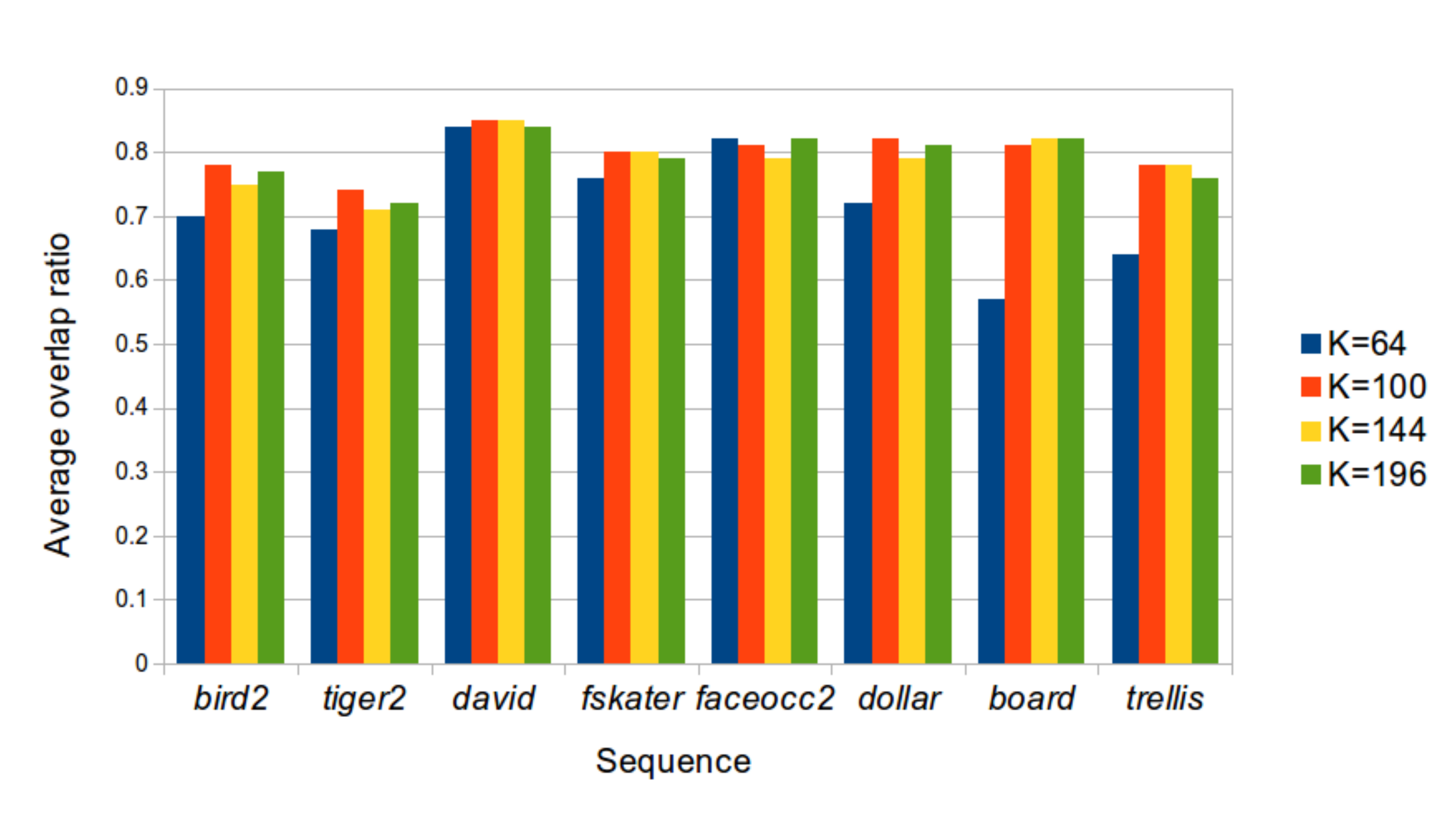}
     \includegraphics[width=0.4985\textwidth]{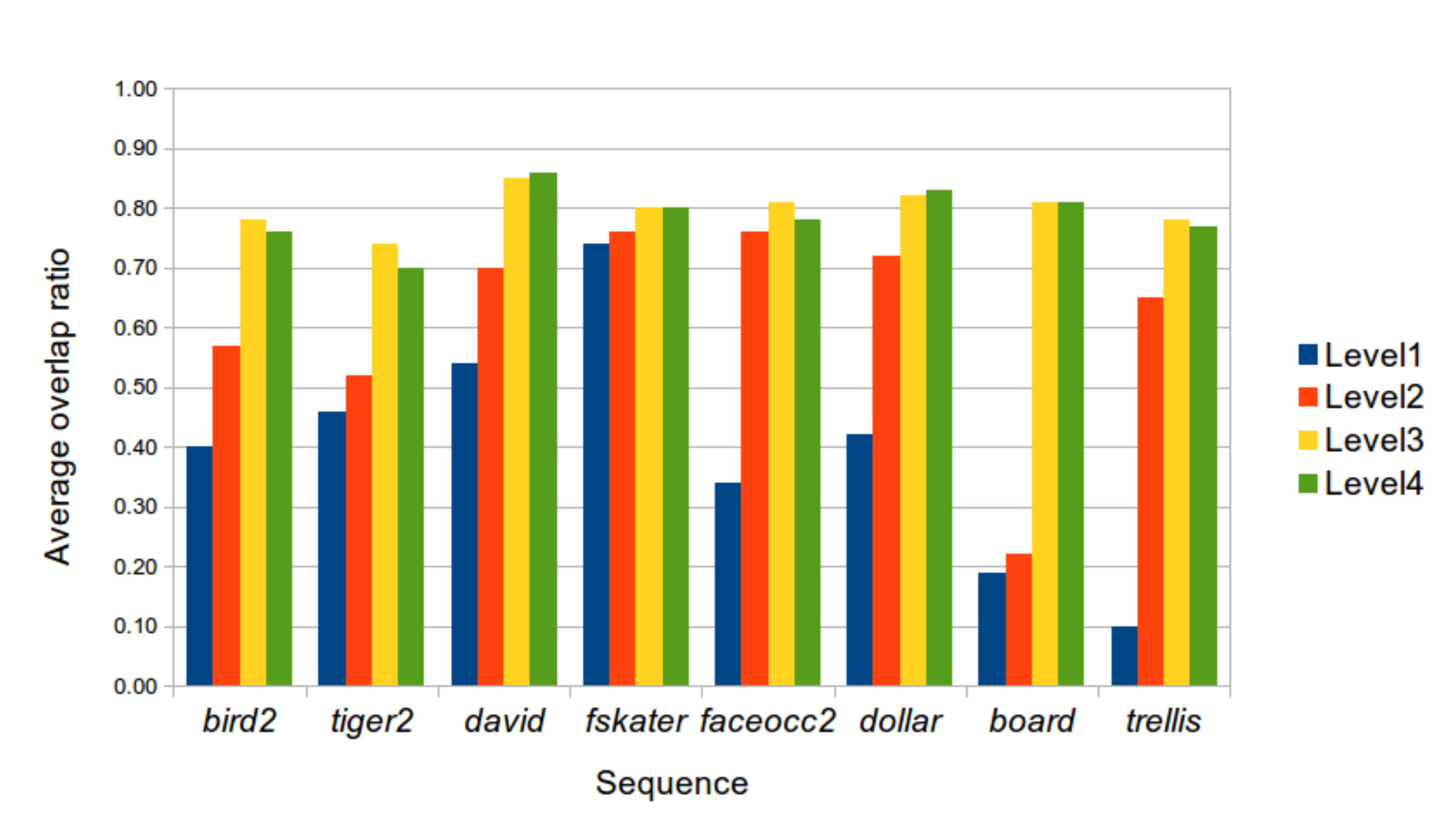}
\caption{Performance comparison of VOR scores
	with different dictionary sizes  (top) and different pooling levels (bottom). }  \label{fig:cbsizeComp}
\end{figure}

\paragraph{Evaluation of different dictionary sizes and pooling levels}
Generally, using larger dictionary and pooling more levels would improve classification
accuracy, which however would inevitably lead to higher dimensional features. Thousands
of dictionary bases are typically used in the image classification task. In the case
of visual tracking, due to the real time limitation, the features can not be too high
dimensional. Fortunately, the image patterns appeared in a tracking scene are relatively
simple, which means that hundreds of dictionary words would be enough to yield good
results. We thus evaluate four dictionary sizes (64, 100, 144, 196) as well as four
pooling levels in terms of VOR scores in figure \ref{fig:cbsizeComp}. It can be seen
that using a dictionary of size 100 greatly promotes the performance on most of the
sequences compared to size 64 except the faceocc2. Further enlarging it does not gain any
significant improvement or even deteriorate the performance. As for the pooling levels,
using more layers improves the tracking accuracy on most of the sequences. However, the
difference gets less notable with the increase of the pooling levels except on the board
sequence where the tracker with two-level pooling features lost the target. When the
pooling level increases to 4, the performance even get worse due to overfitting. Based on
these observations, we choose a dictionary size of 100 and 3-level pooling as a compromise
for accuracy and efficiency to report the tracking results in Table \ref{tab:overlapScore}
and \ref{tab:CLE}.

 \paragraph{Comparing with other features in Struck}
 To further demonstrate the strength of the learned features, we incorporate the feature learning into the Struck framework and compare with three other types of features originally used in \cite{struck}, which are raw pixel, Haar and histogram features. Linear kernel is used here for evaluation. All the other settings are the same with \cite{struck} for all the sequences. Table \ref{tab:Struck_feat} reports the average VORs and CLEs on eight sequences.
 As can be can be observed, different hand-crafted features perform well in  particular scenarios as they capture different information of the tracking scene, while as the learned feature achieves the overall best performance and outperforms its counterparts significantly. In conclusion, the features learned in a principled fashion is superior than the traditional hand-crafted features in tracking tasks.

\begin{table} \center
\resizebox{1\linewidth}{!}
{
\begin{tabular}{ r | c  c | c  c | c  c | c  c }
\hline
\multirow{2}{*}{{{Sequence}}} &\multicolumn{2}{c|}{{ST}} &\multicolumn{2}{c|}{{Raw}}
&\multicolumn{2}{c|}{{Haar}} &\multicolumn{2}{c}{{Histogram}} \\
&{VOR} &{CLE} &{VOR} &{CLE} &{VOR} &{CLE} &{VOR} &{CLE}\\
\hline
	\emph{david} & \textbf{0.82} &\textbf{7.1} &0.38 &51.3 &0.46 &45.5 & 0.70 &14.3\\
	\emph{girl} & 0.78 &12.2 & 0.75 &13.8 & \textbf{0.80} &\textbf{10.4} & 0.28 &67.4 \\
	\emph{faceocc1} & 0.81 &11.4 & 0.82 &11.0 & \textbf{0.84} &\textbf{9.5} & 0.79 &13.1  \\
	\emph{faceocc2} & 0.79 &\textbf{6.9} & 0.75 &12.2 & \textbf{0.81} &9.4 & 0.68 &16.6 \\
	\emph{coke11} & \textbf{0.75} &\textbf{4.3} & 0.66 &6.5 & 0.64 &6.5 & 0.56  &9.7\\
	\emph{tiger1} & \textbf{0.76} &\textbf{4.9} & 0.68 &7.6 & 0.32 &37.7 & 0.72  &5.6  \\
	\emph{tiger2} & \textbf{0.71} & \textbf{6.4} & 0.48 &11.6 & 0.56 &13.2 & 0.61  &9.4  \\
	\emph{sylvester} & \textbf{0.77} &7.0 & 0.69 &9.2 & 0.62 &11.7 & 0.76 &\textbf{6.8} \\
	\hline
	\emph{average} &\bf{0.77} &\textbf{7.5}  &0.65  &15.4 &0.63 &18.0 &0.64 &17.9\\
	\hline
\end{tabular}
}
\caption{Performance comparison (VORs and CLEs)
	of different features using Struck with linear kernel.
		The last row shows the averaged results over all sequences.
		The best results are bold faced.} \label{tab:Struck_feat}
\end{table}

 \subsection{Discussions on dictionary update}
\begin{figure}[]
\center
     \includegraphics[width=0.3415\textwidth]{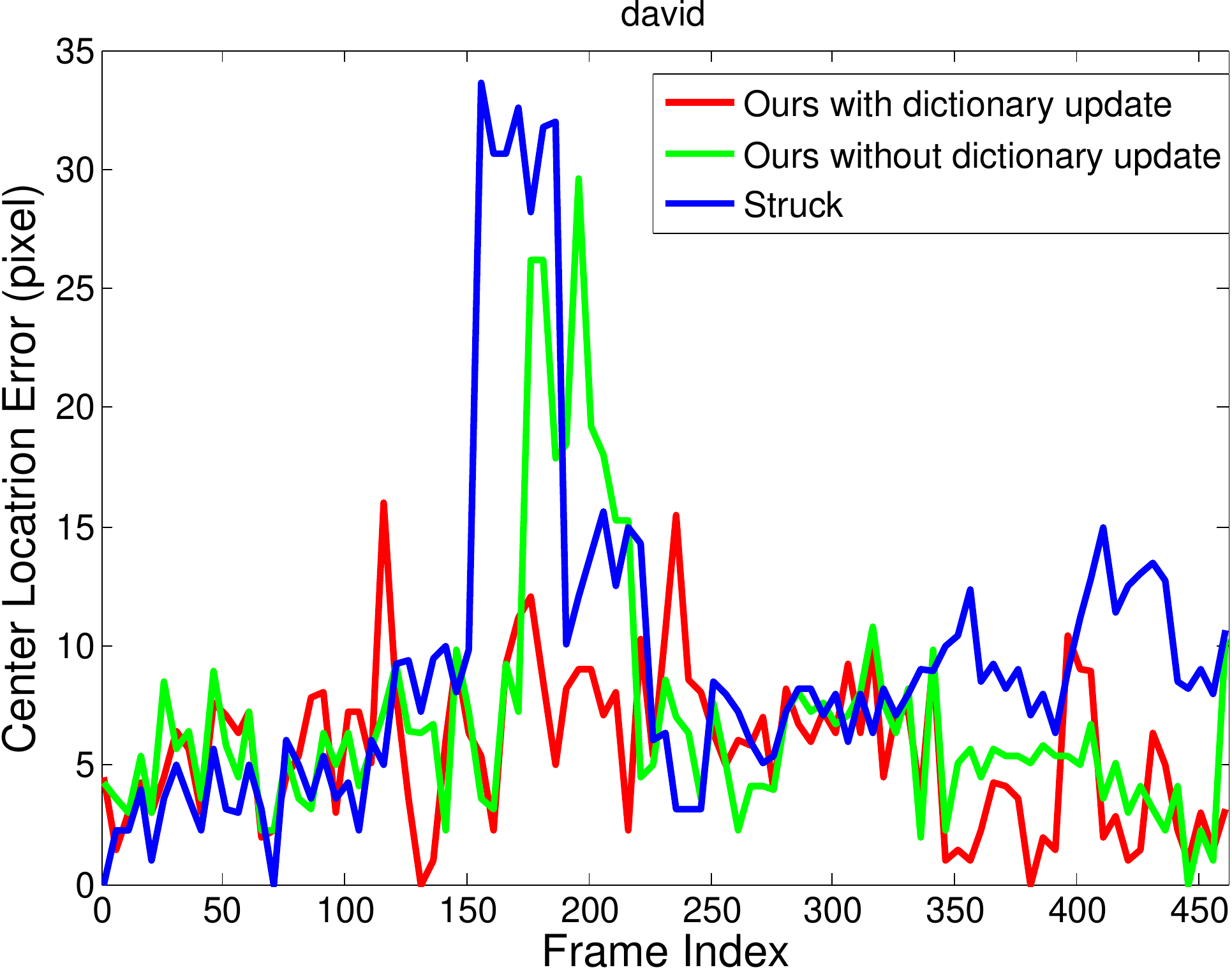}
     \\
\vspace{0.2cm}
	 \includegraphics[width=0.3415\textwidth]{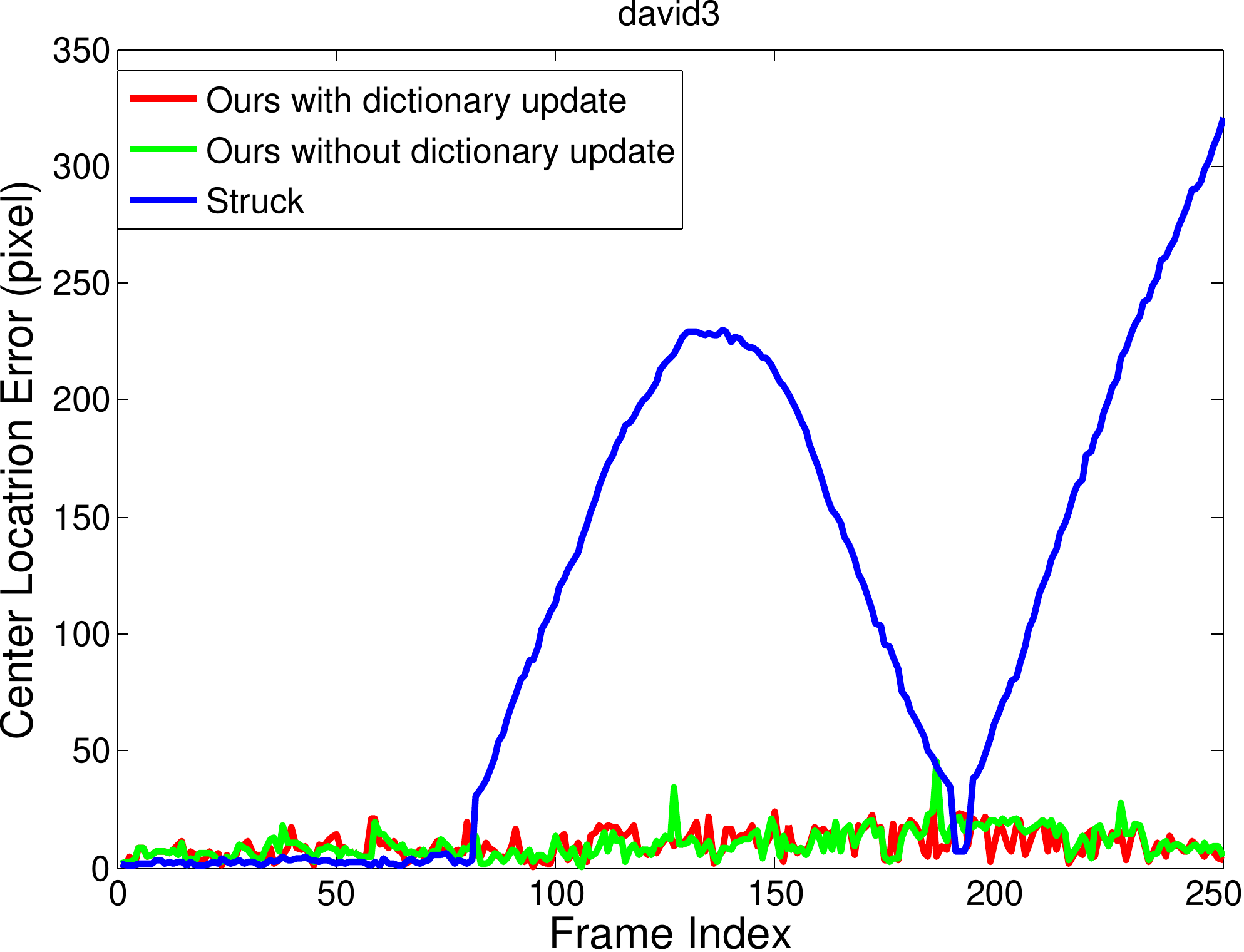}  \\
\vspace{0.2cm}
     \includegraphics[width=0.3415\textwidth]{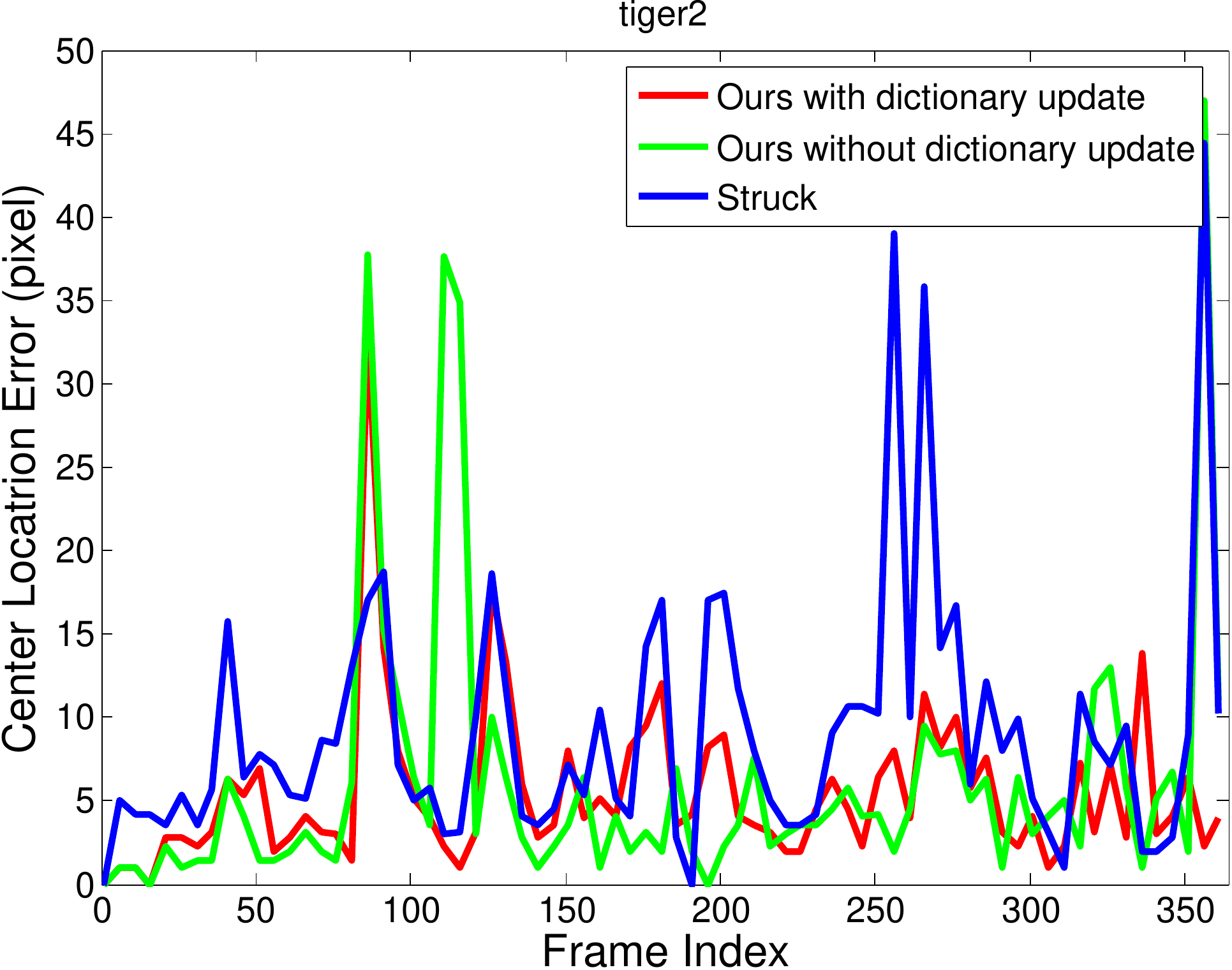}
\\
\vspace{0.2cm}
	\includegraphics[width=0.3415\textwidth]{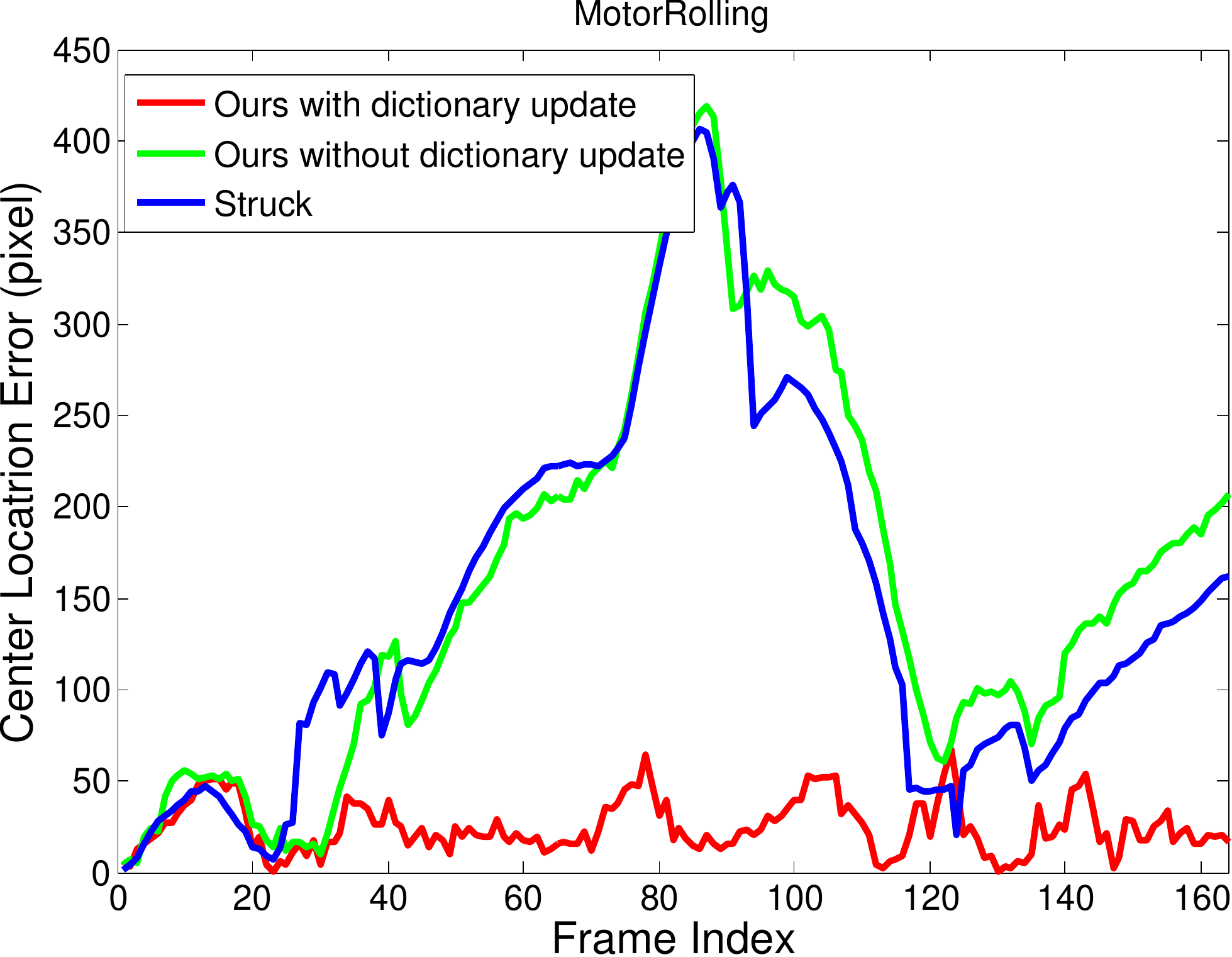}

\caption{The center location error plots of our tracker with and without dictionary update compared with Struck on four sequences. }  \label{fig:CLEplot}
\end{figure}

 From the reported results, we can see that using the dictionary simply learned from the patches extracted in the first frame yields surprisingly satisfactory results, almost as good as its counterpart with updating scheme. We conjecture that this advantage comes from the fact that most of the tracking sequences consists of relatively simple  patterns. Even with various changes, the scenes are similar. To demonstrate the effectiveness of the dictionary update scheme proposed here, we find a sequence with drastic scene changes as well as $360^\circ$ in-plane rotation of the object, which is motorRolling. Without update, our tracker lost the target at the frame 38 and yield a final VOR of 0.11 with the CLE 160.3. While equipped with the dictionary updating scheme, it tracks the target during the whole process giving an average VOR of 0.49 with the CLE 24.9, although not accurate enough due to the severe variations of the object appearance. Figure \ref{fig:CLEplot} shows the center location error plots of our method both with and without dictionary update compared to Struck on four sequences.

\begin{figure*}
\center
     \includegraphics[width=0.161\textwidth, height=0.12\textwidth]{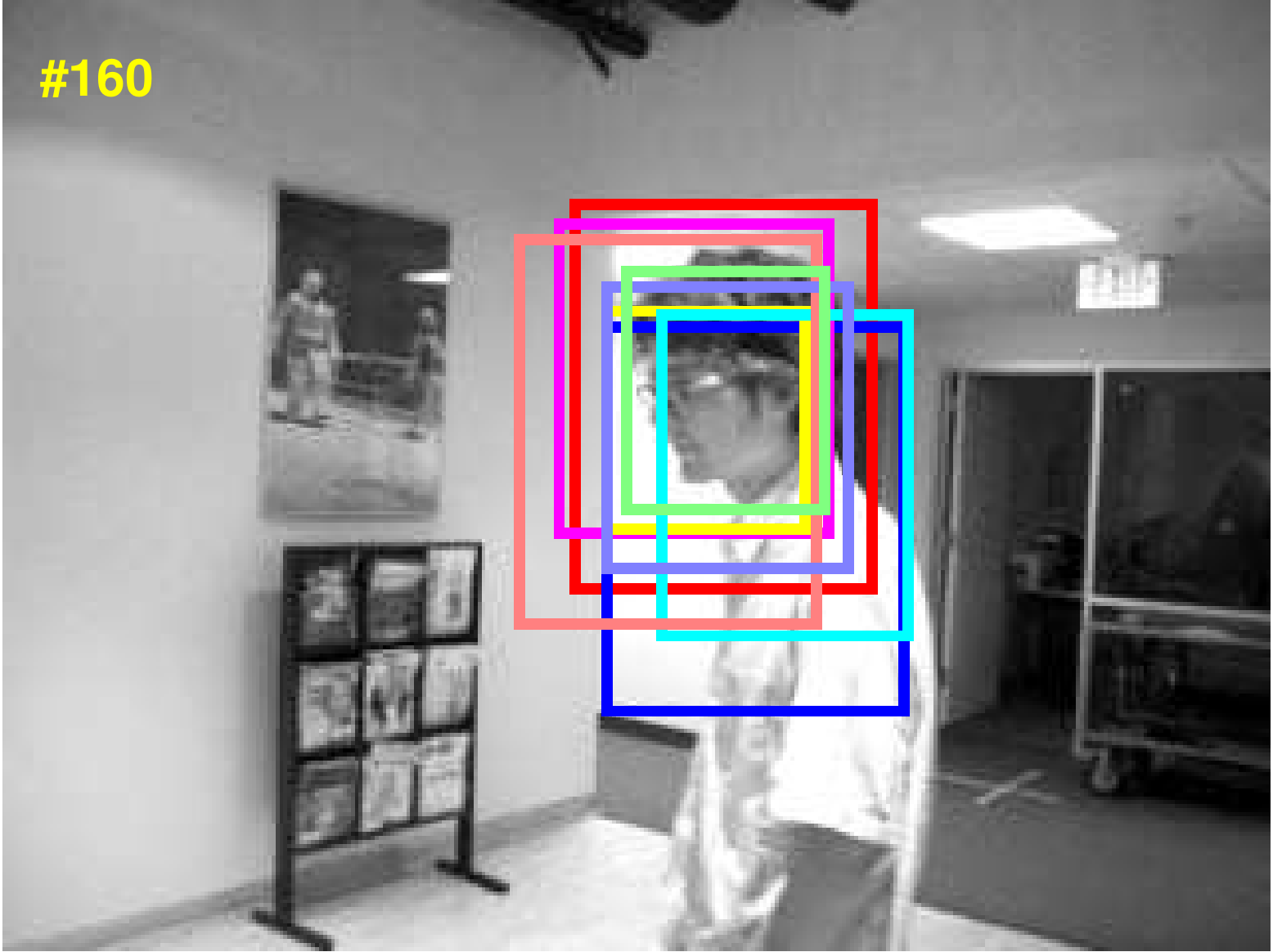}
	 \includegraphics[width=0.161\textwidth, height=0.12\textwidth]{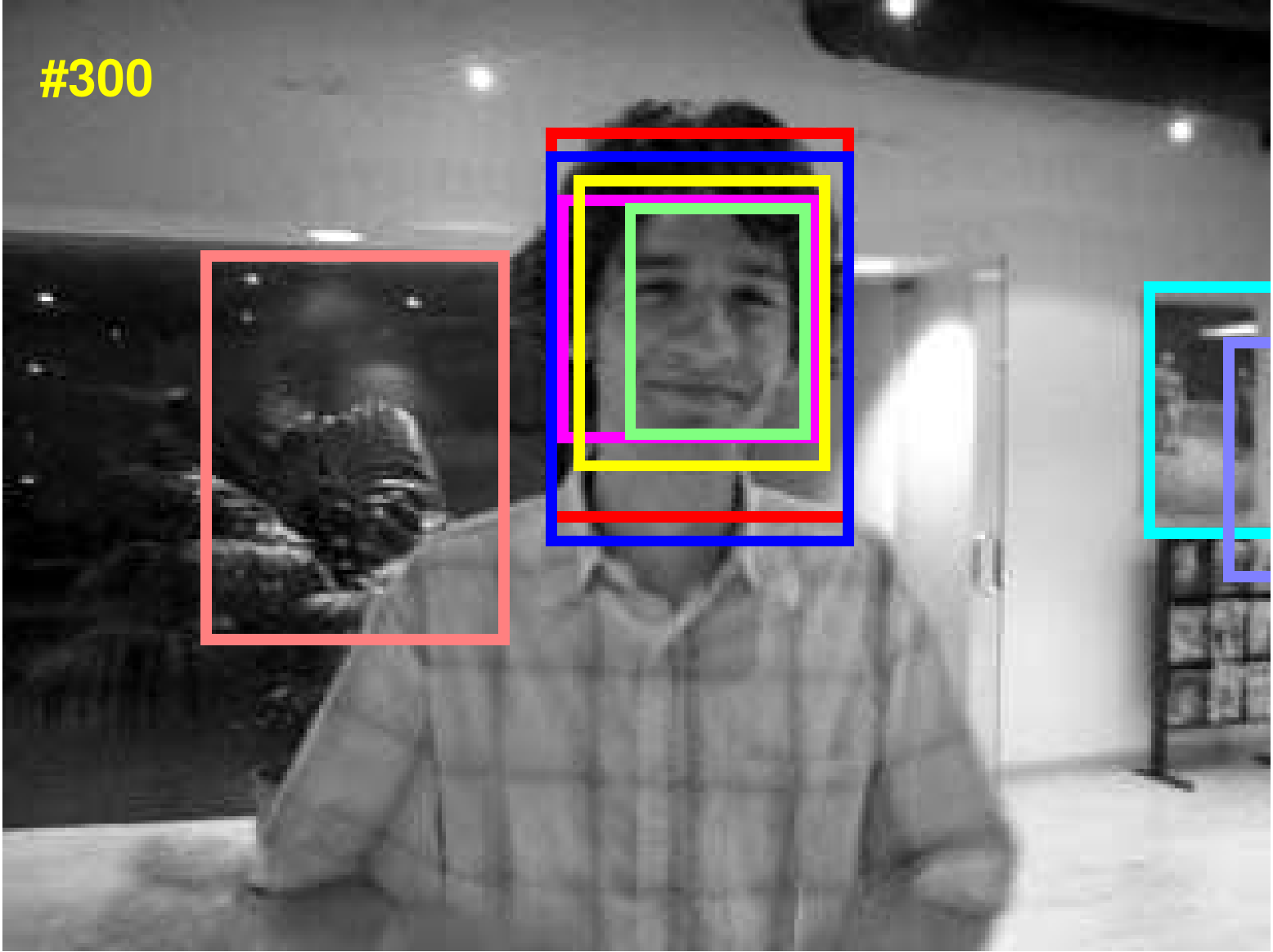}
	 \includegraphics[width=0.161\textwidth, height=0.12\textwidth]{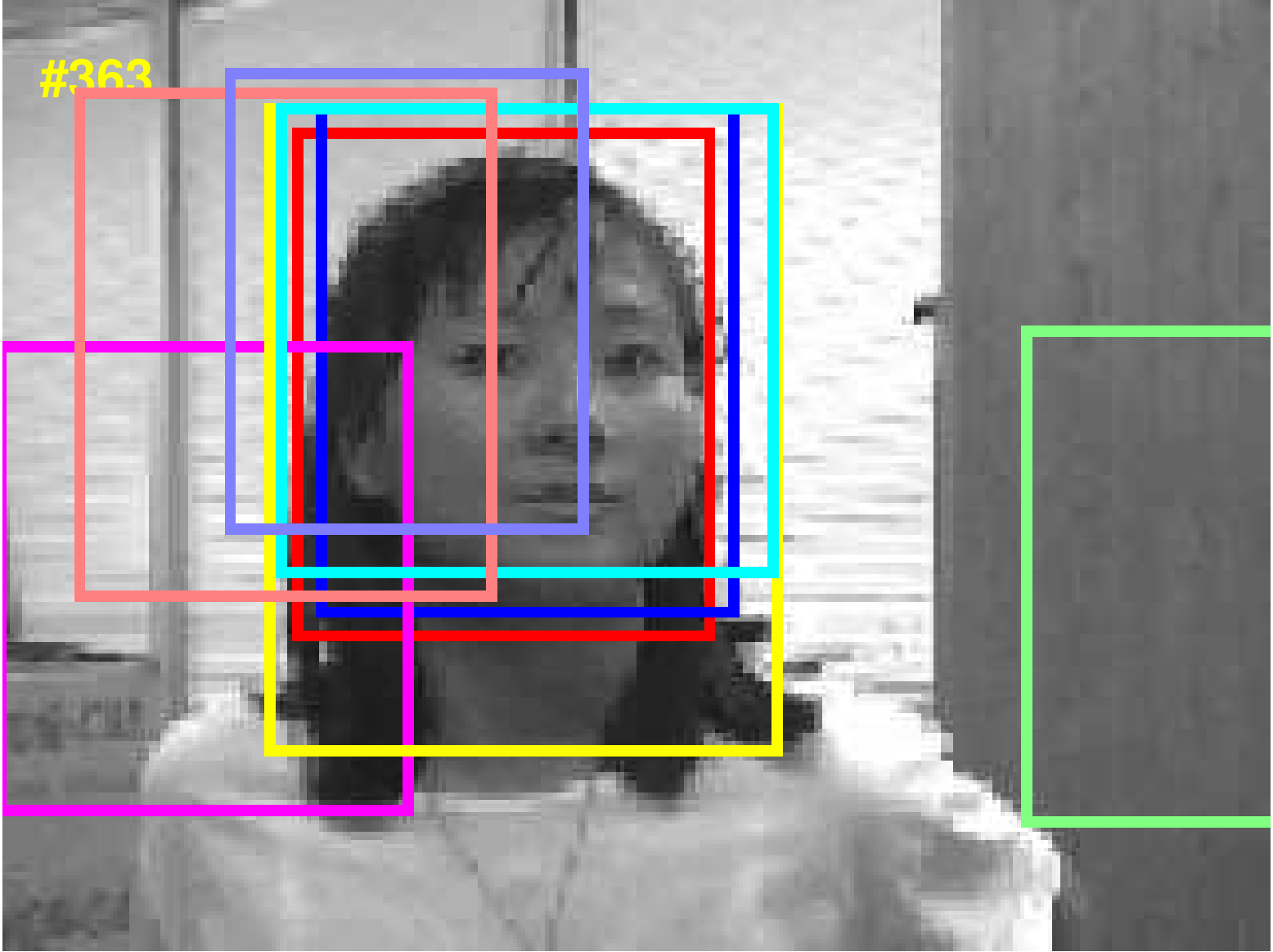}
	 \includegraphics[width=0.161\textwidth, height=0.12\textwidth]{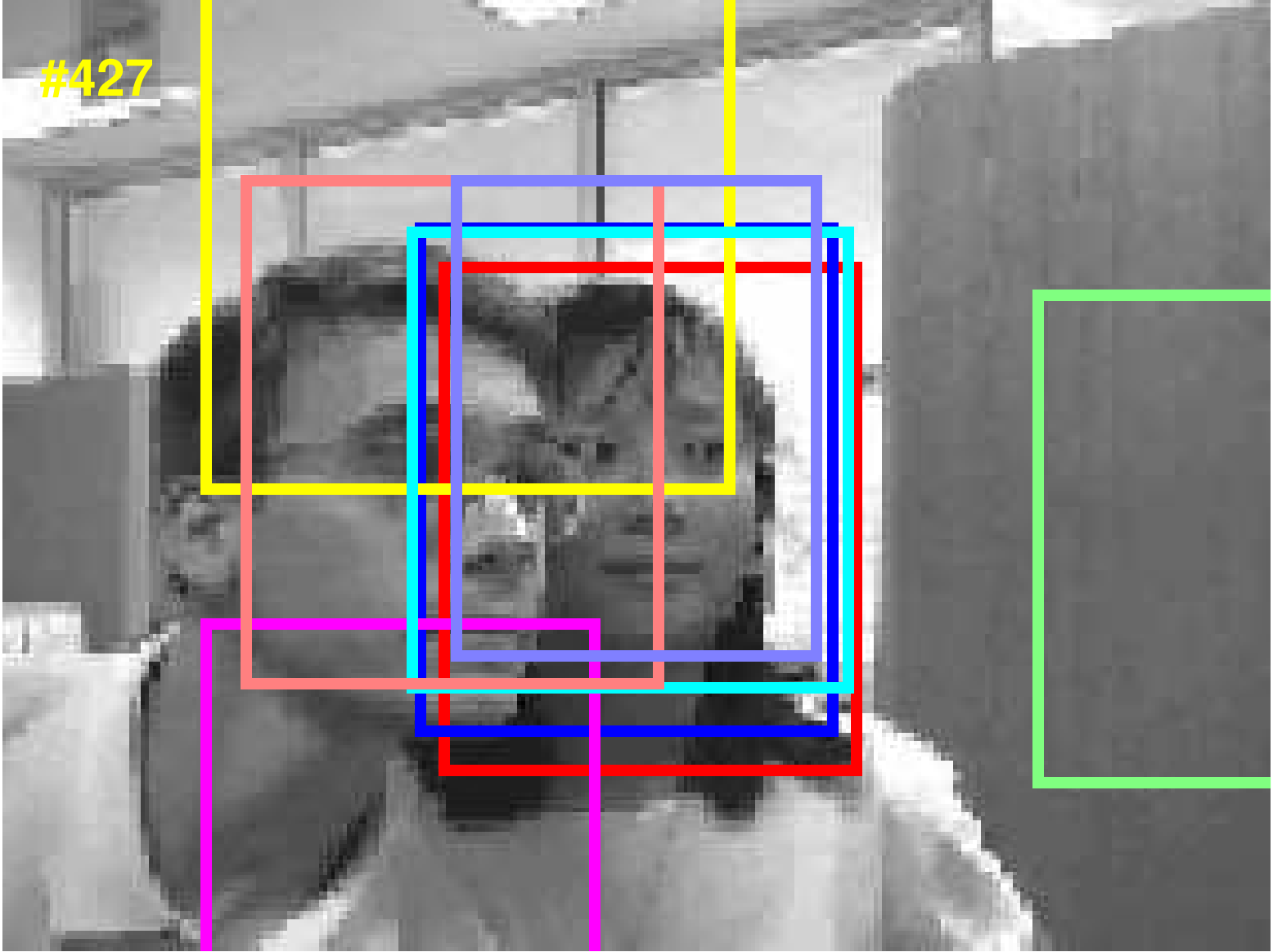}
     \includegraphics[width=0.161\textwidth, height=0.12\textwidth]{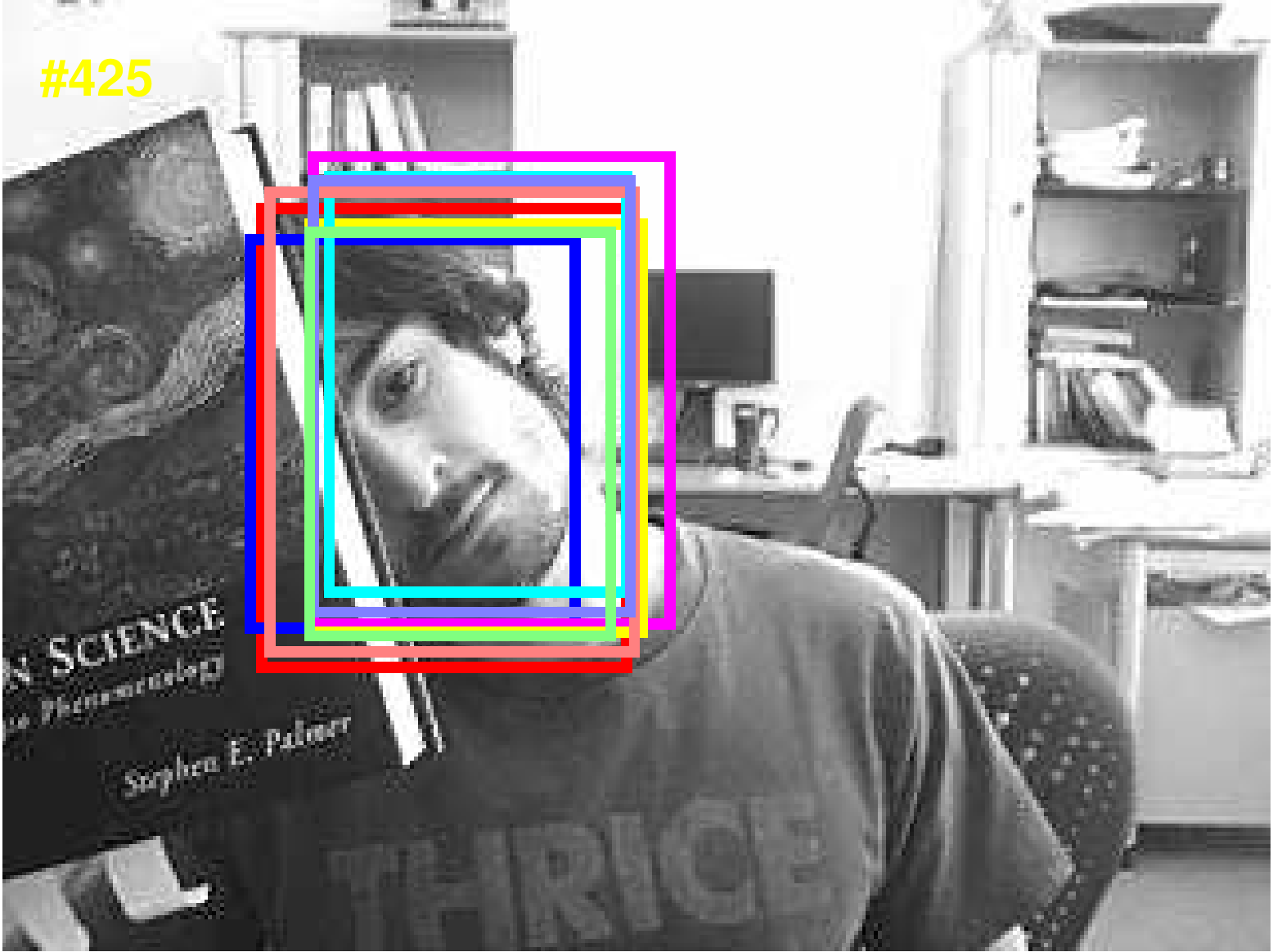}
     \includegraphics[width=0.161\textwidth, height=0.12\textwidth]{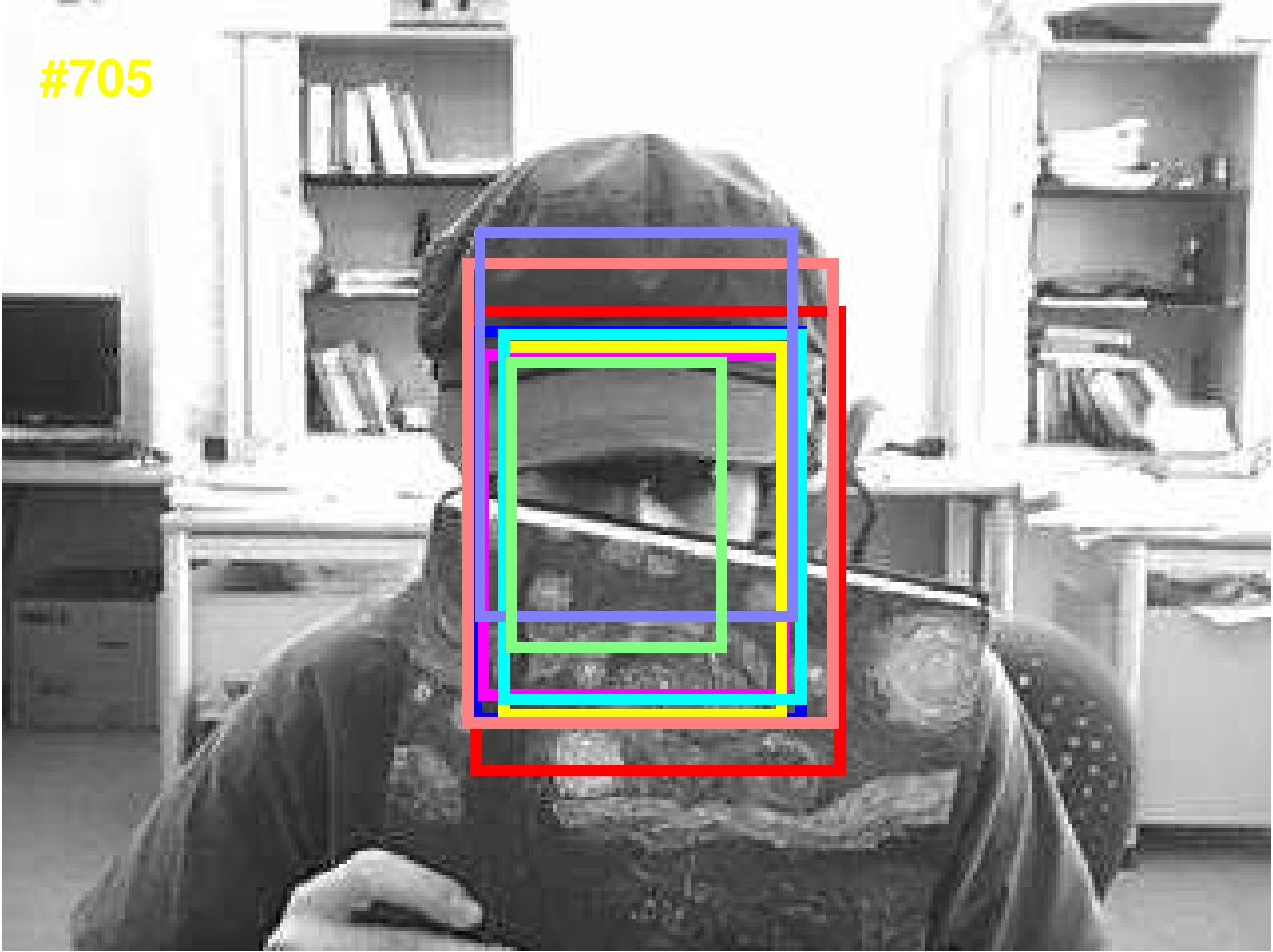} \\

     \includegraphics[width=0.161\textwidth, height=0.12\textwidth]{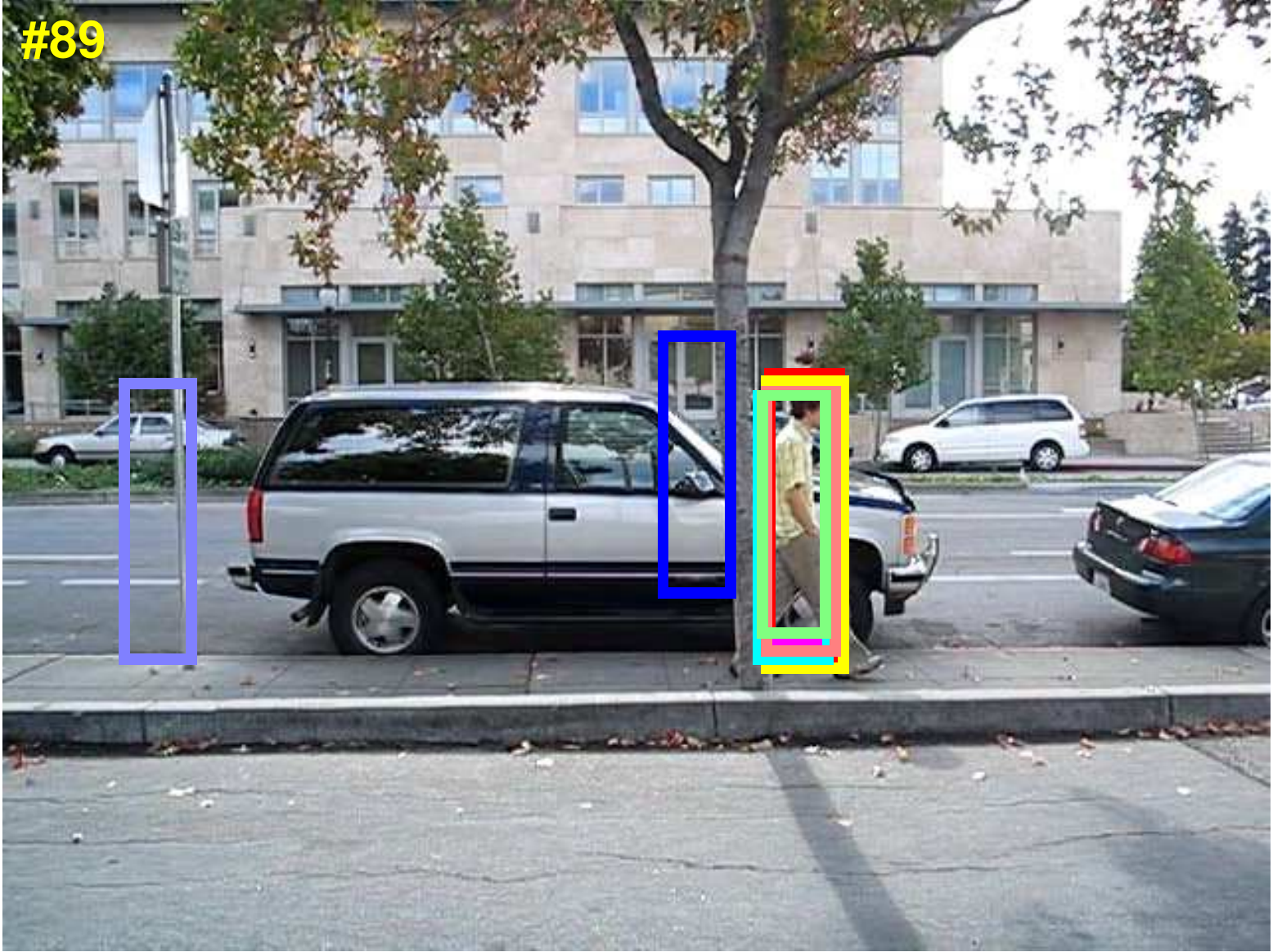}
	 \includegraphics[width=0.161\textwidth, height=0.12\textwidth]{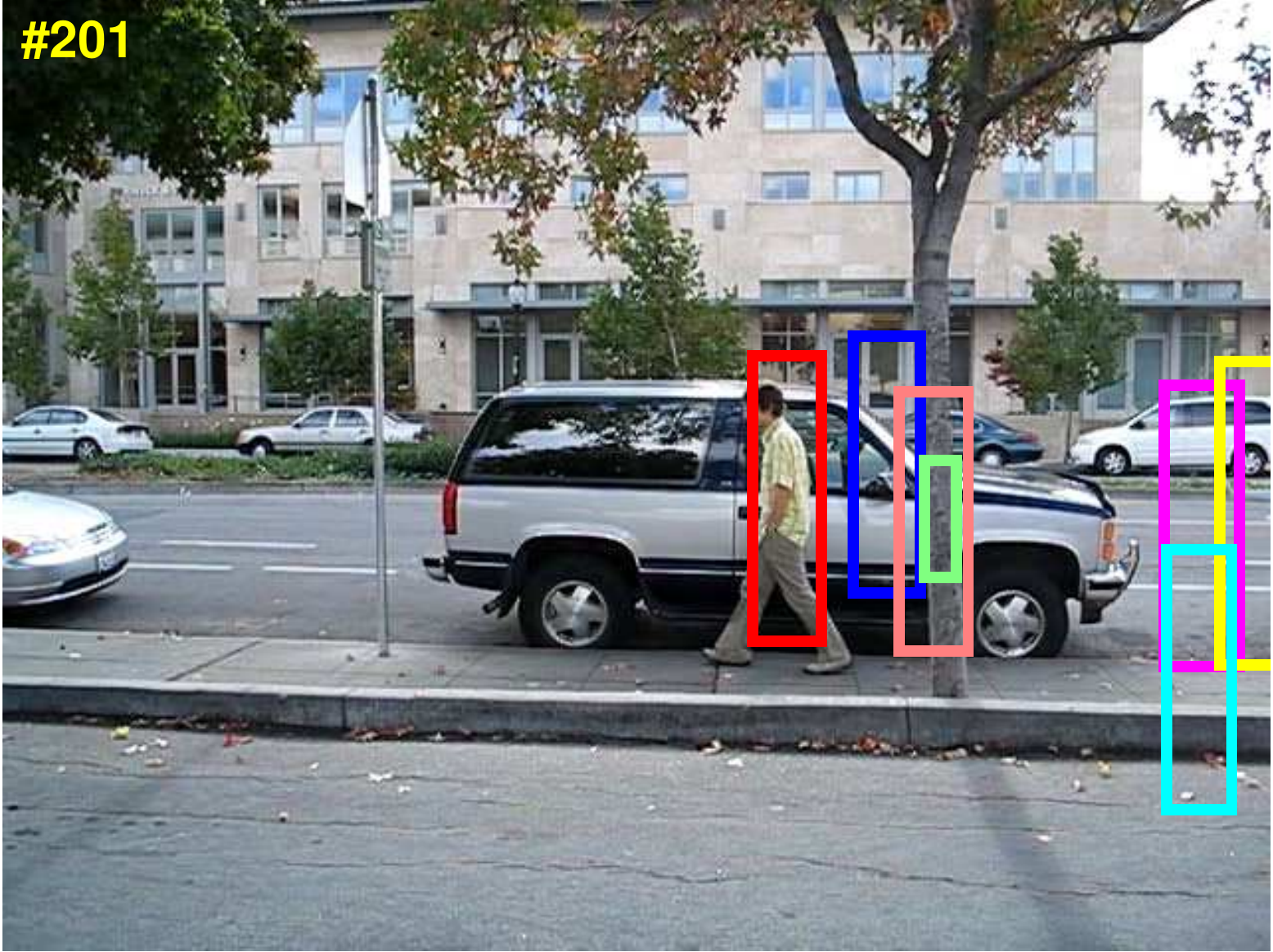}
	 \includegraphics[width=0.161\textwidth, height=0.12\textwidth]{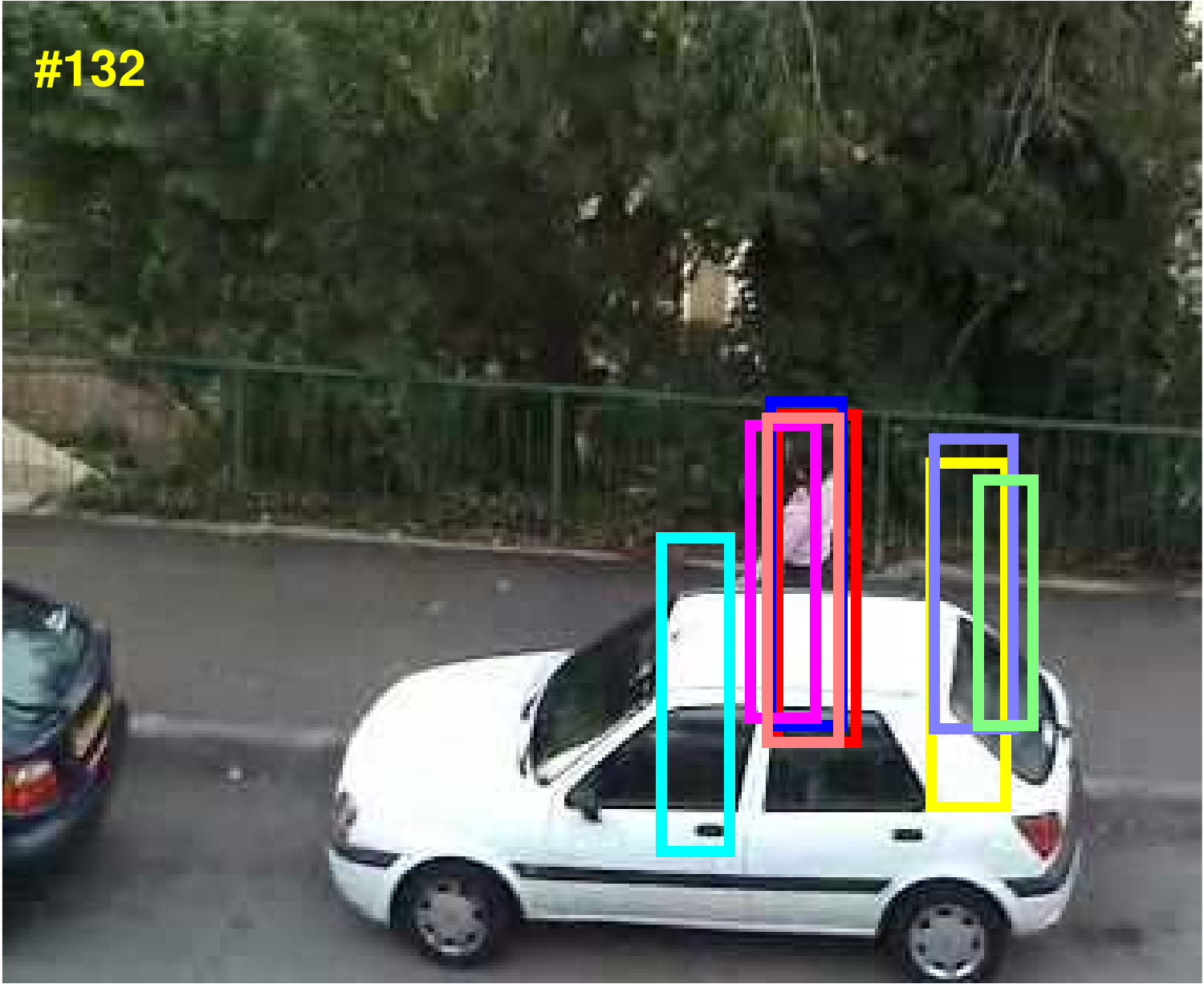}
	 \includegraphics[width=0.161\textwidth, height=0.12\textwidth]{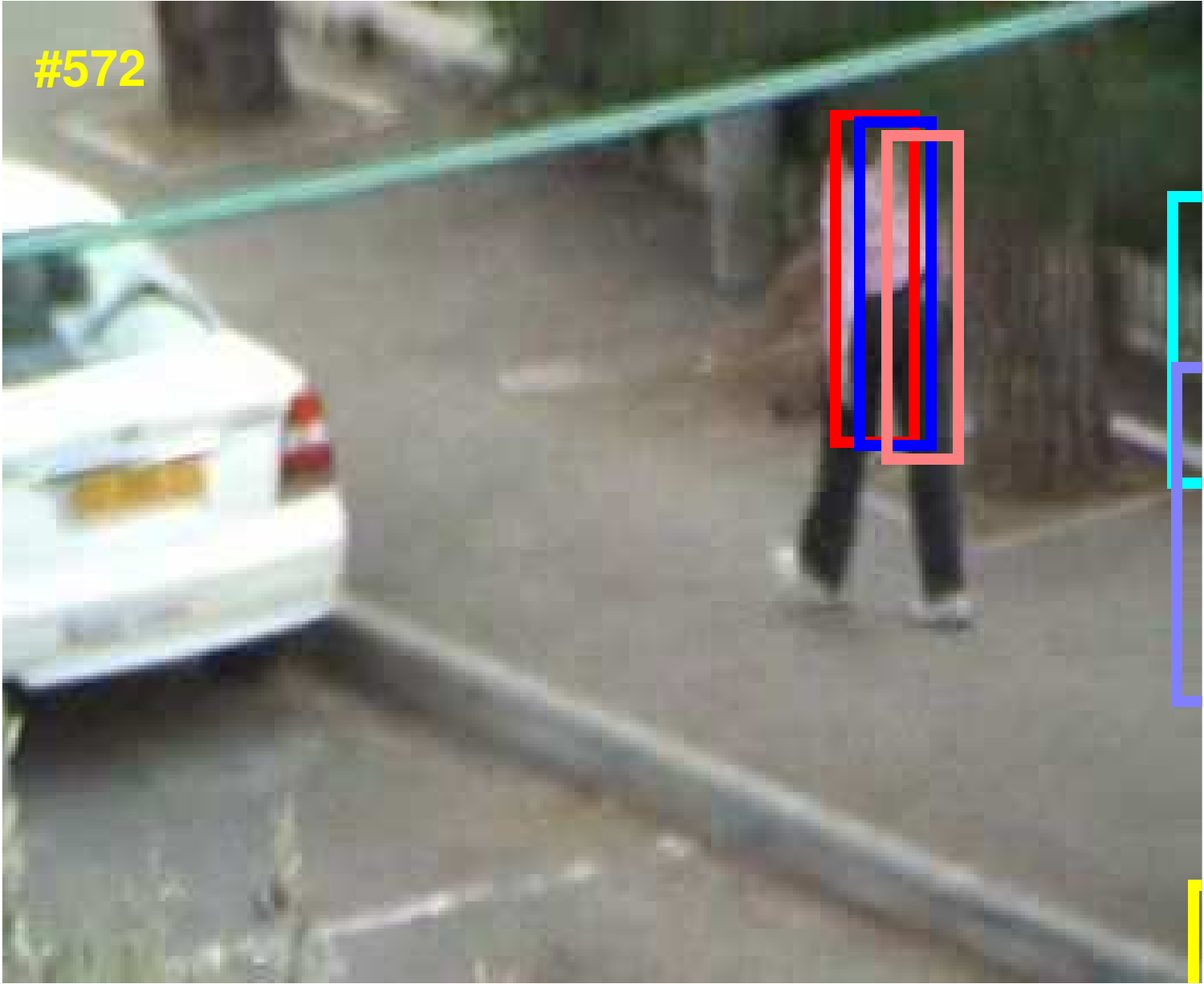}
	  \includegraphics[width=0.161\textwidth, height=0.12\textwidth]{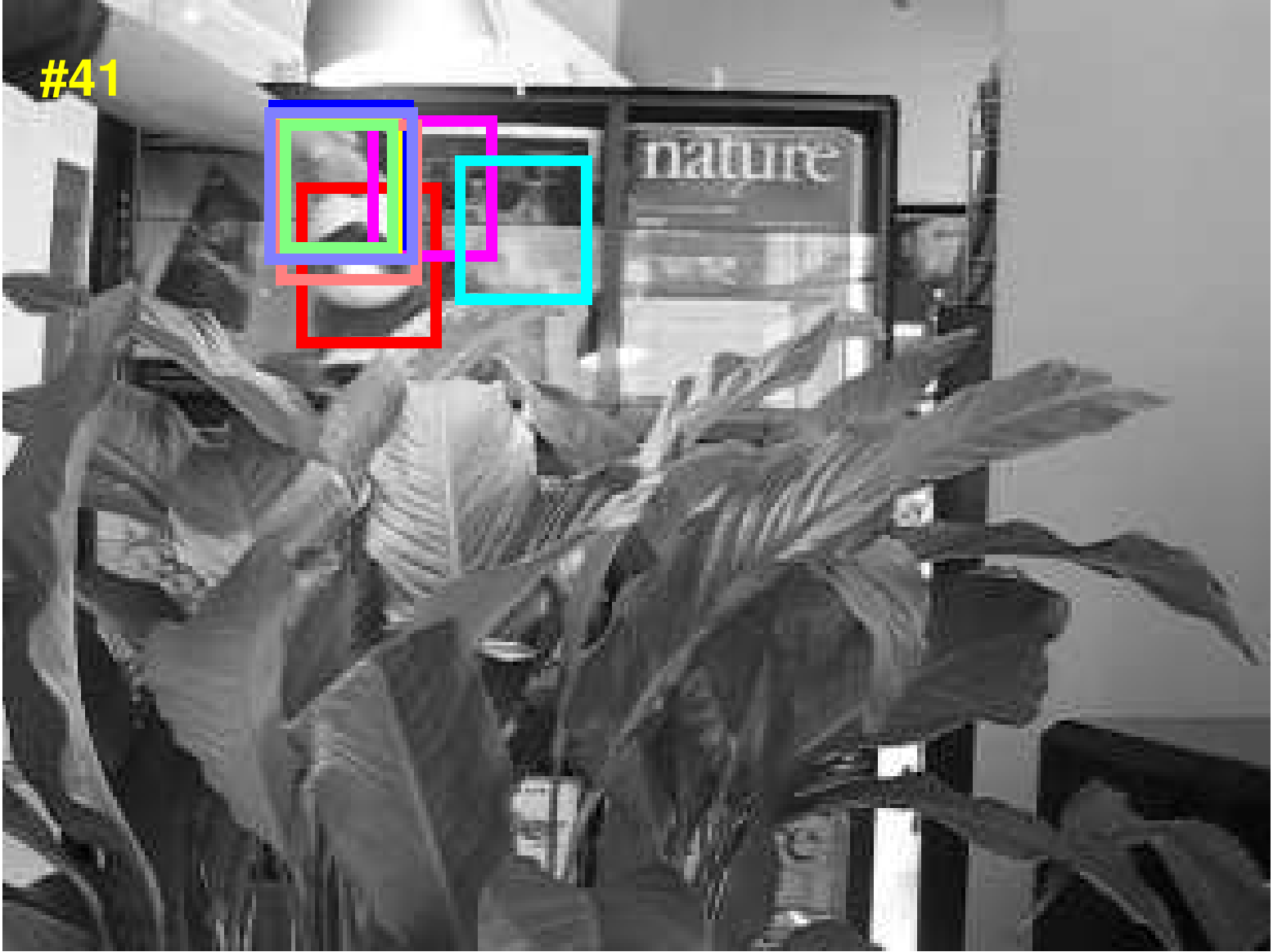}
	 \includegraphics[width=0.161\textwidth, height=0.12\textwidth]{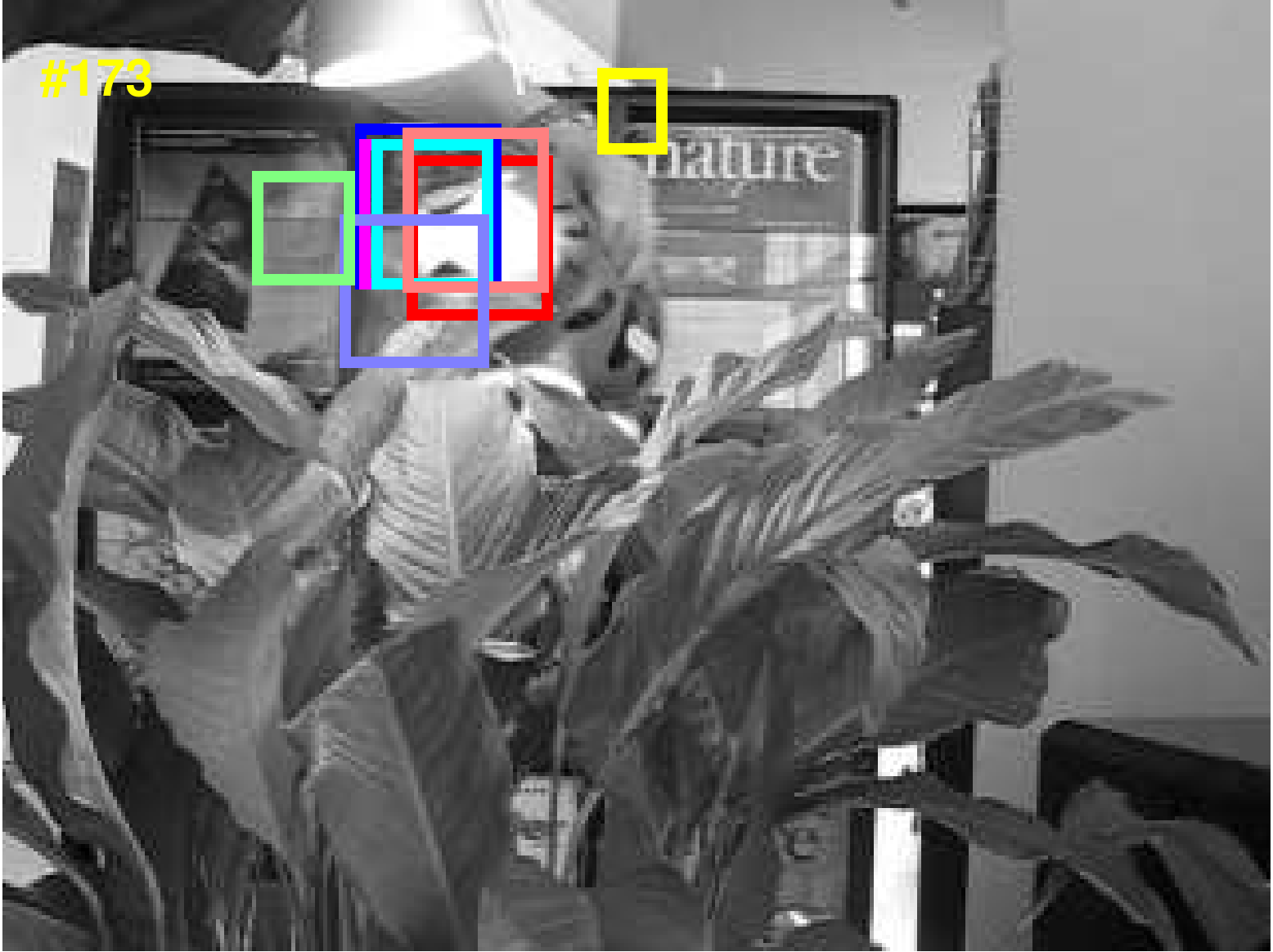} \\

	 \includegraphics[width=0.161\textwidth, height=0.12\textwidth]{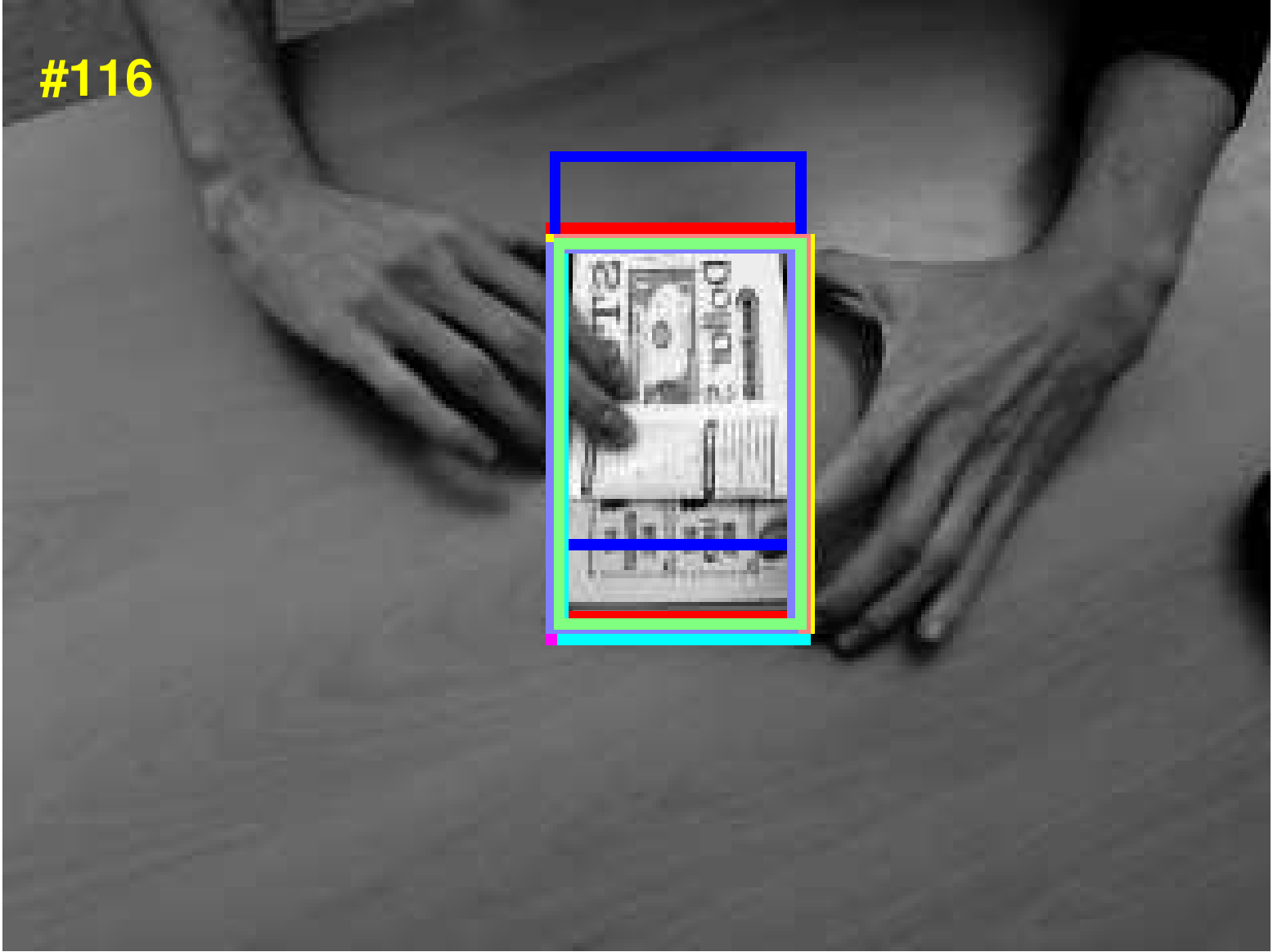}
	 \includegraphics[width=0.161\textwidth, height=0.12\textwidth]{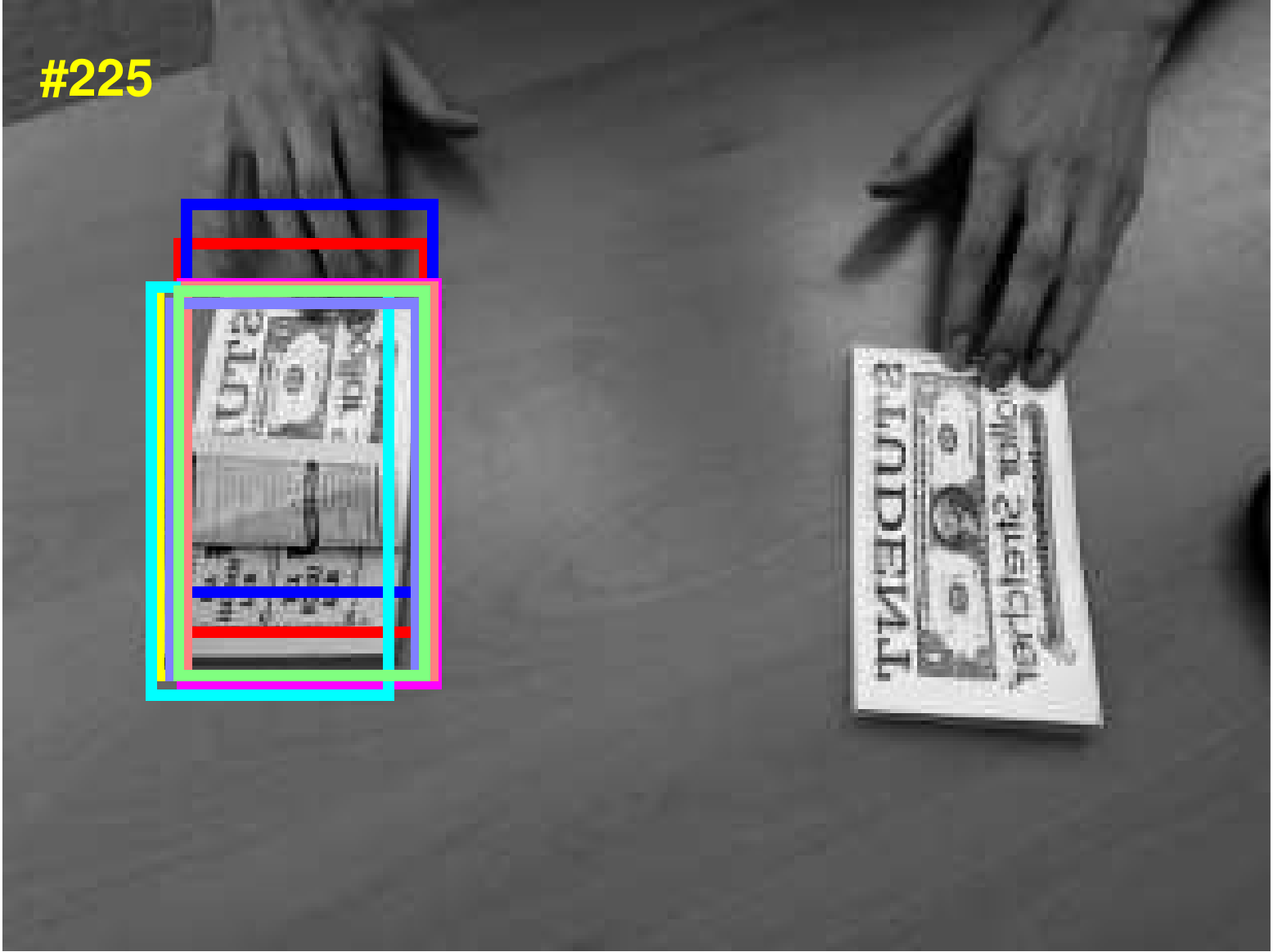}
	 \includegraphics[width=0.161\textwidth, height=0.12\textwidth]{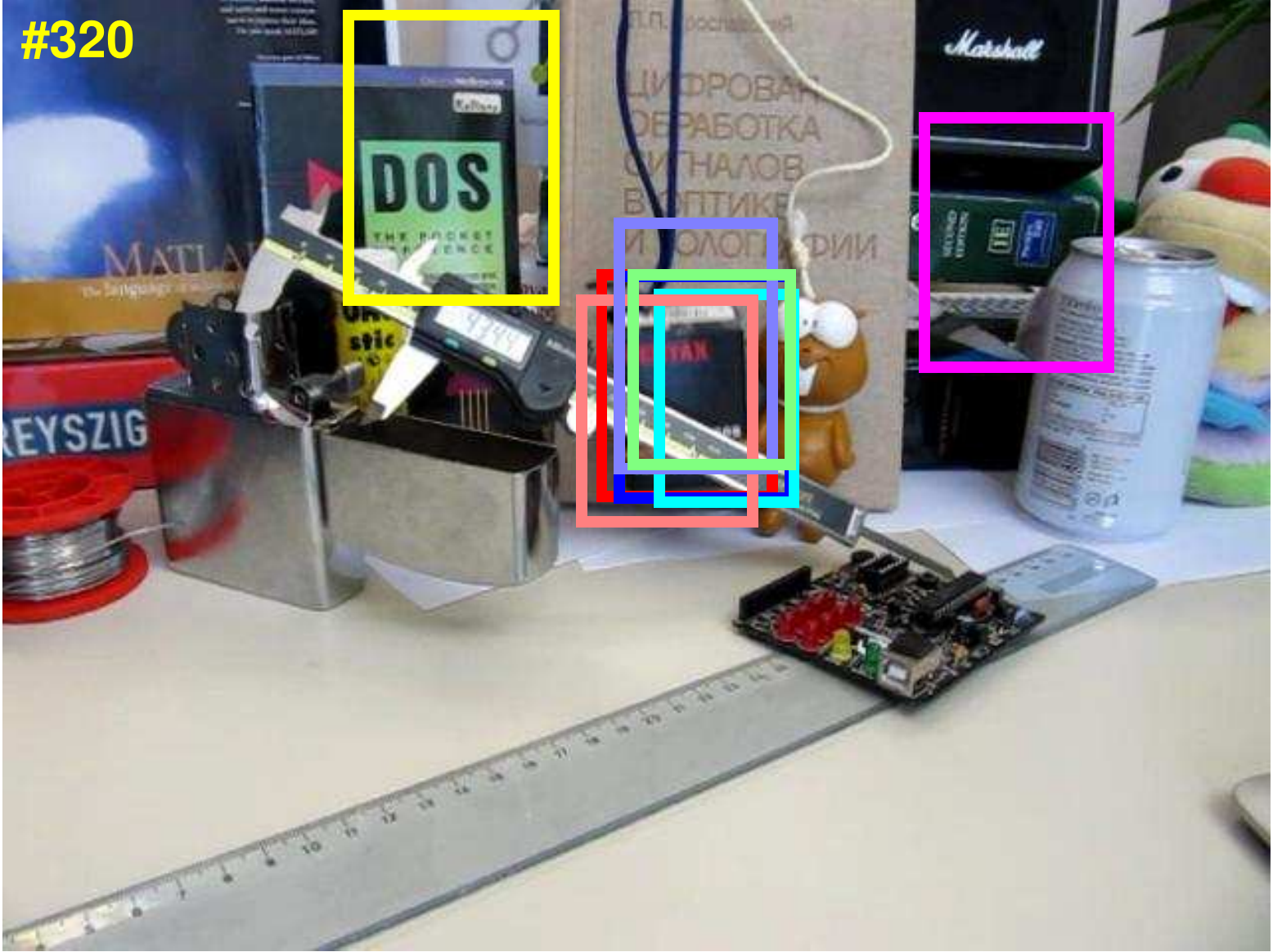}
	 \includegraphics[width=0.161\textwidth, height=0.12\textwidth]{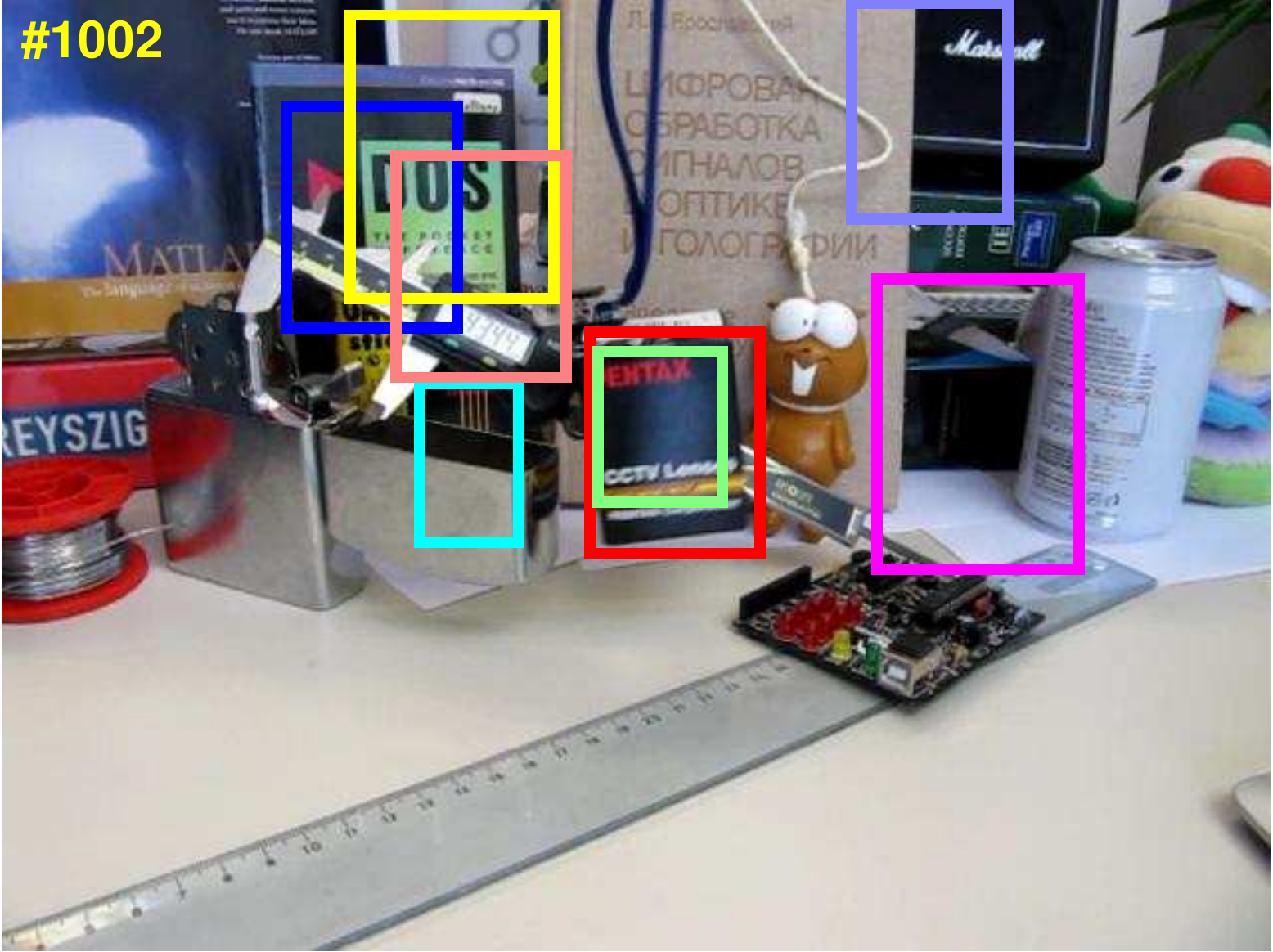}
	 \includegraphics[width=0.161\textwidth, height=0.12\textwidth]{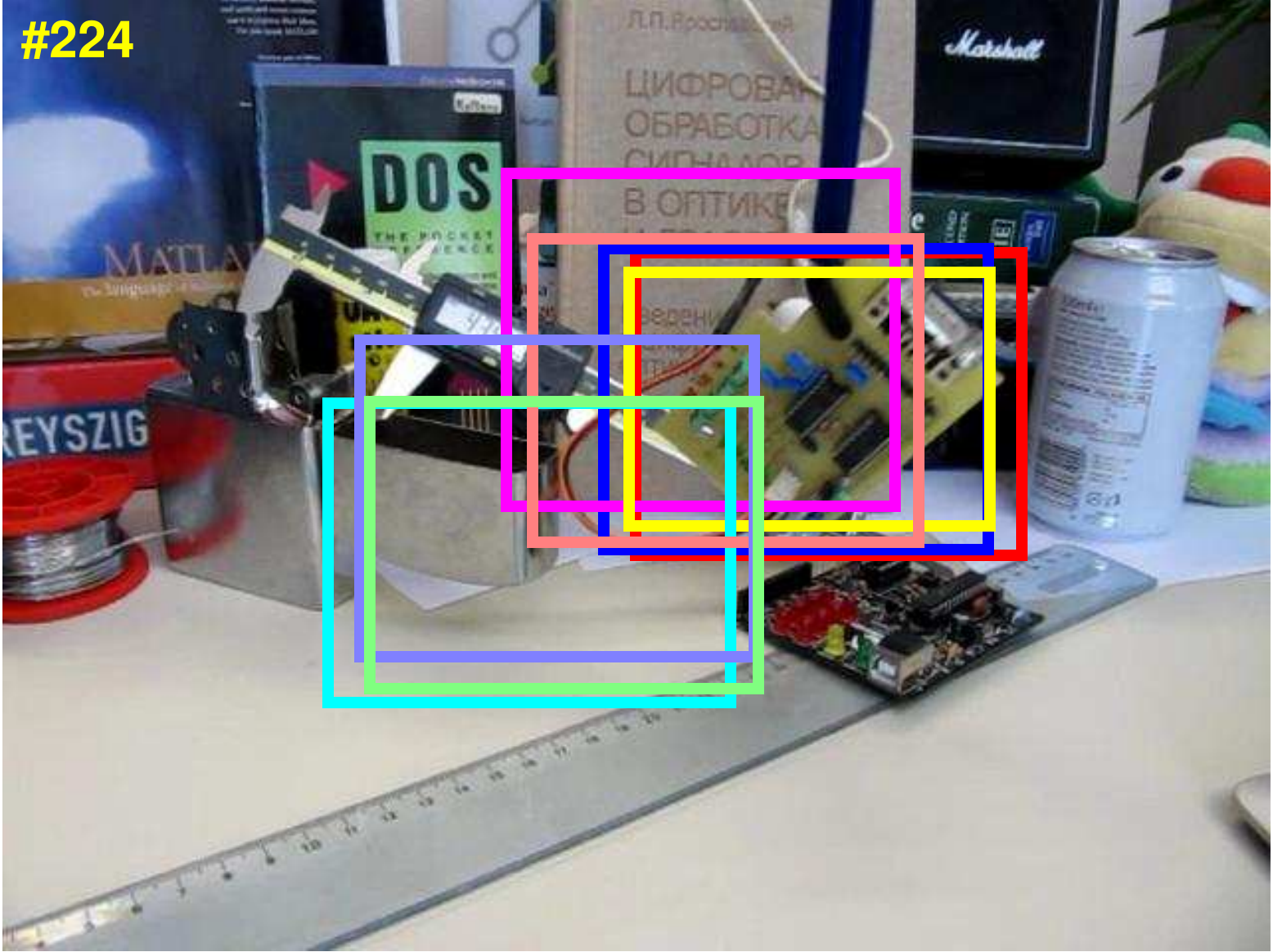}
	 \includegraphics[width=0.161\textwidth, height=0.12\textwidth]{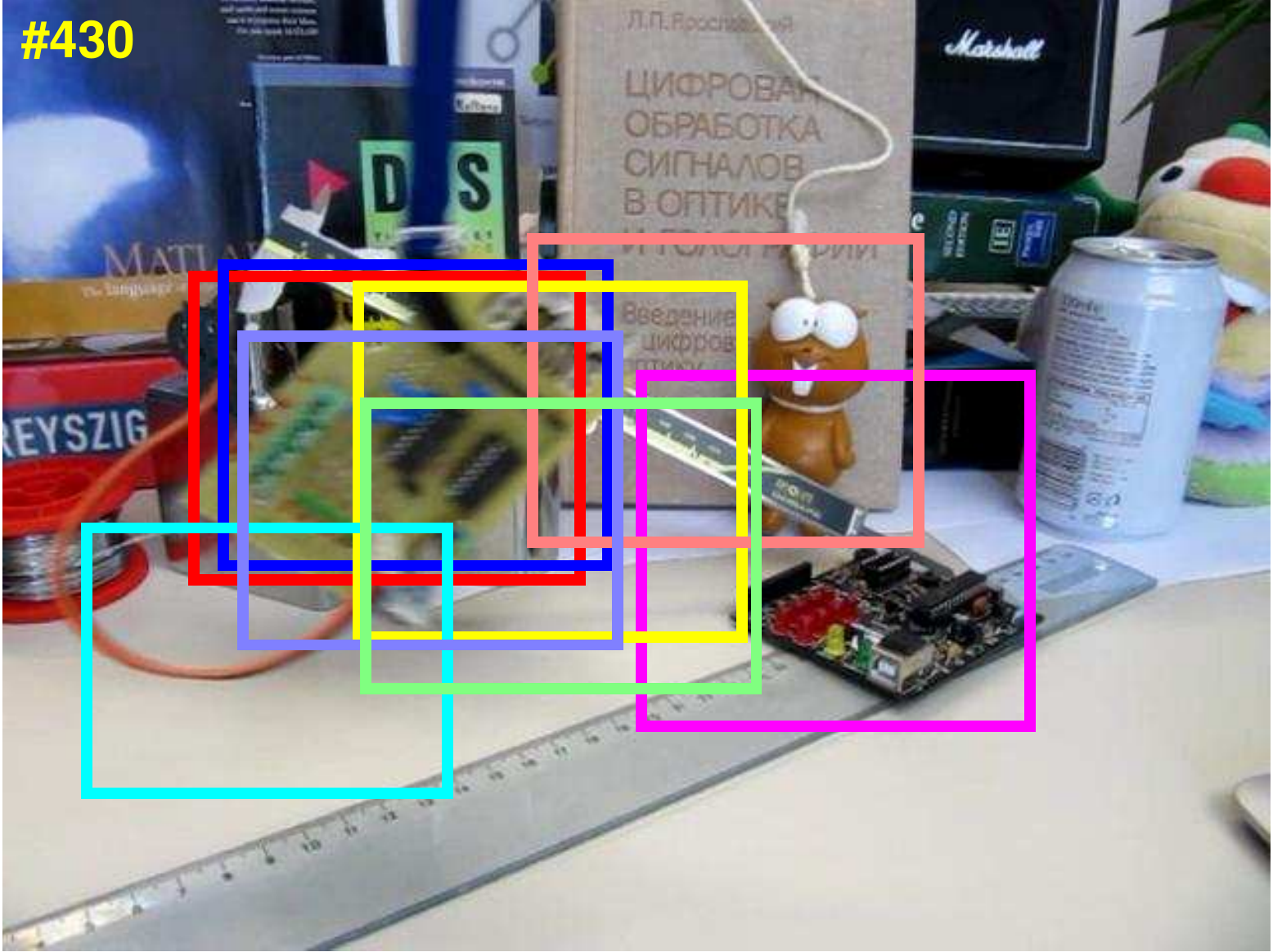} \\

	 \includegraphics[width=0.161\textwidth, height=0.12\textwidth]{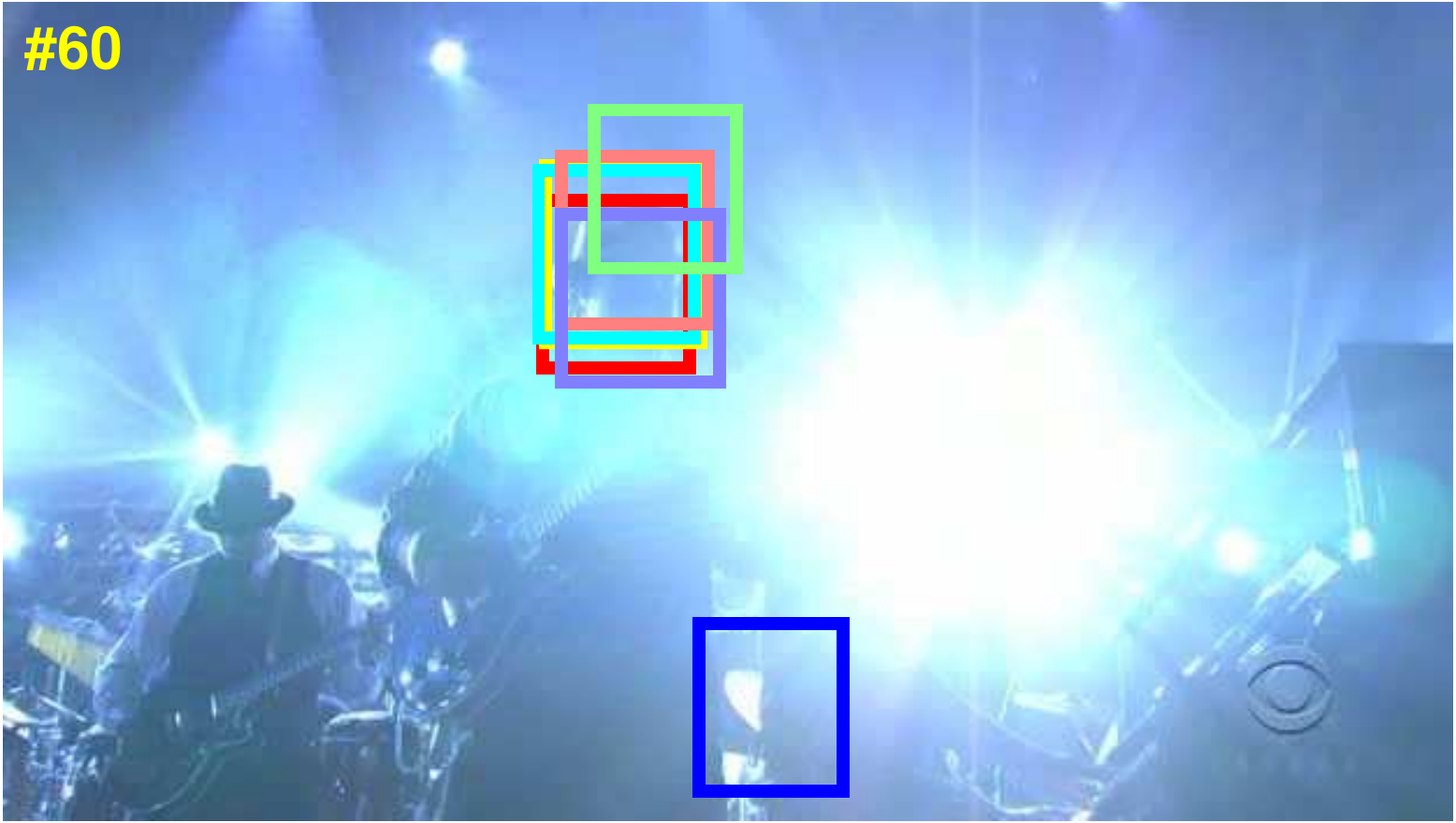}
	 \includegraphics[width=0.161\textwidth, height=0.12\textwidth]{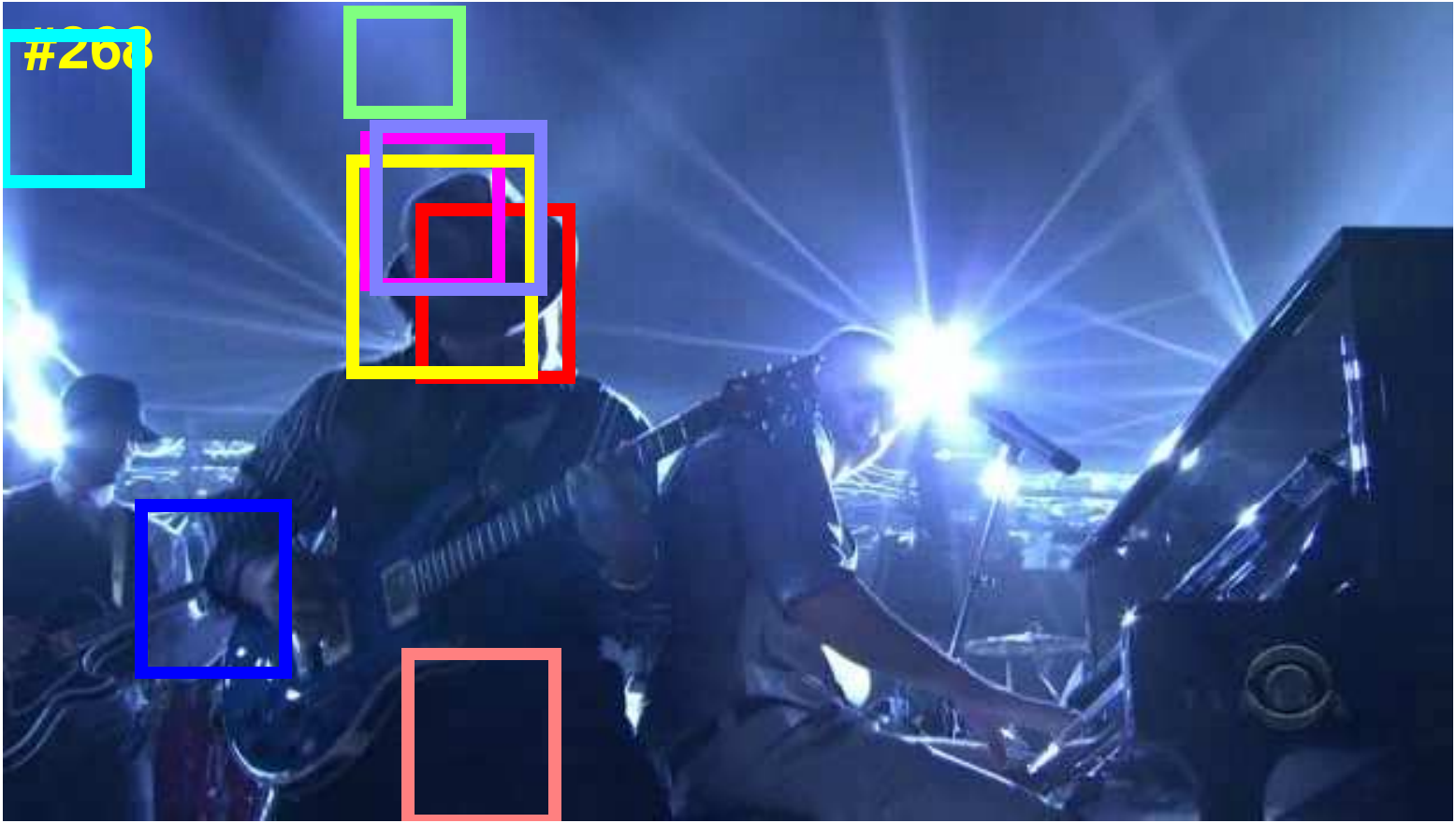}
	 \includegraphics[width=0.161\textwidth, height=0.12\textwidth]{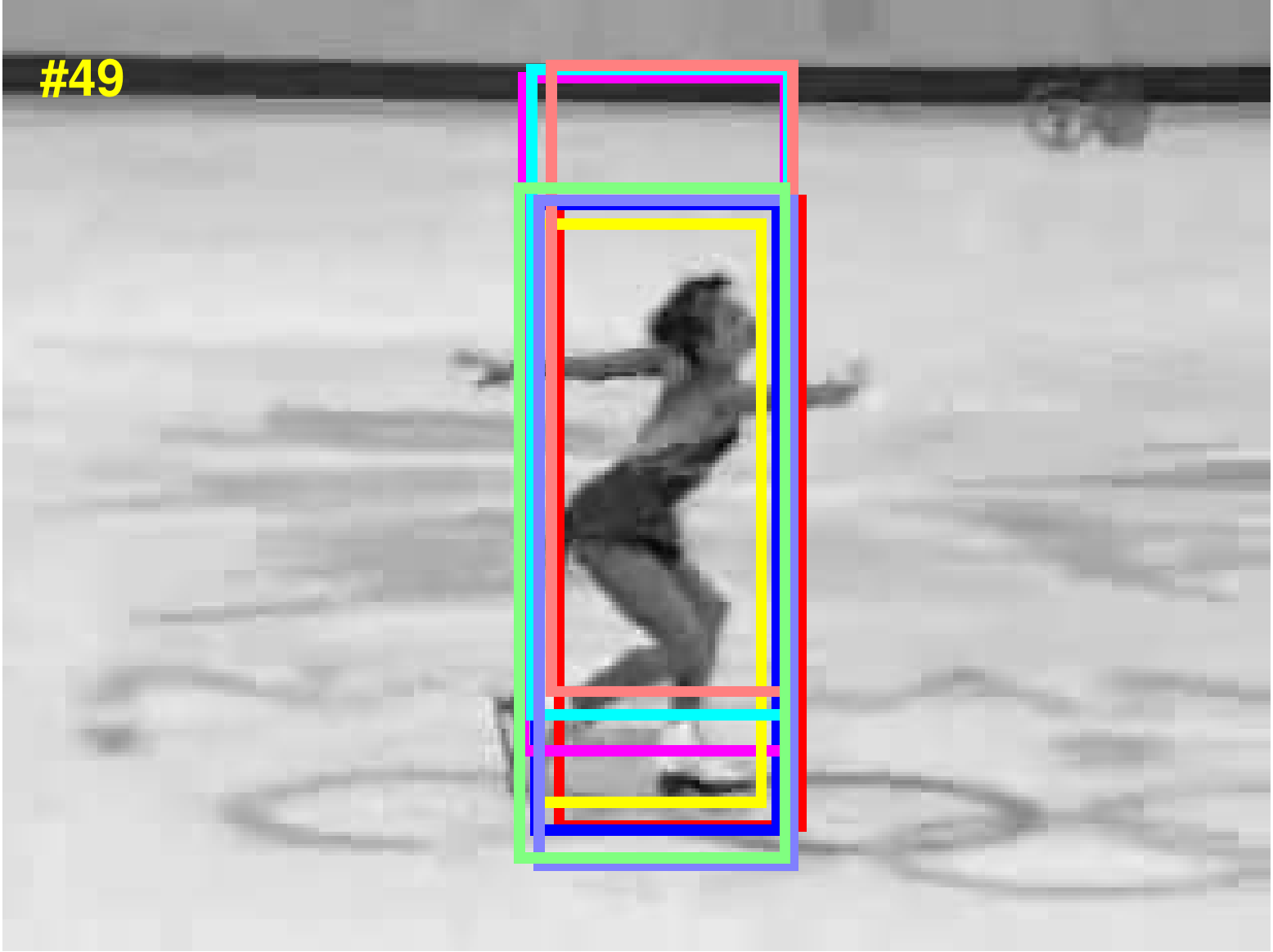}
	 \includegraphics[width=0.161\textwidth, height=0.12\textwidth]{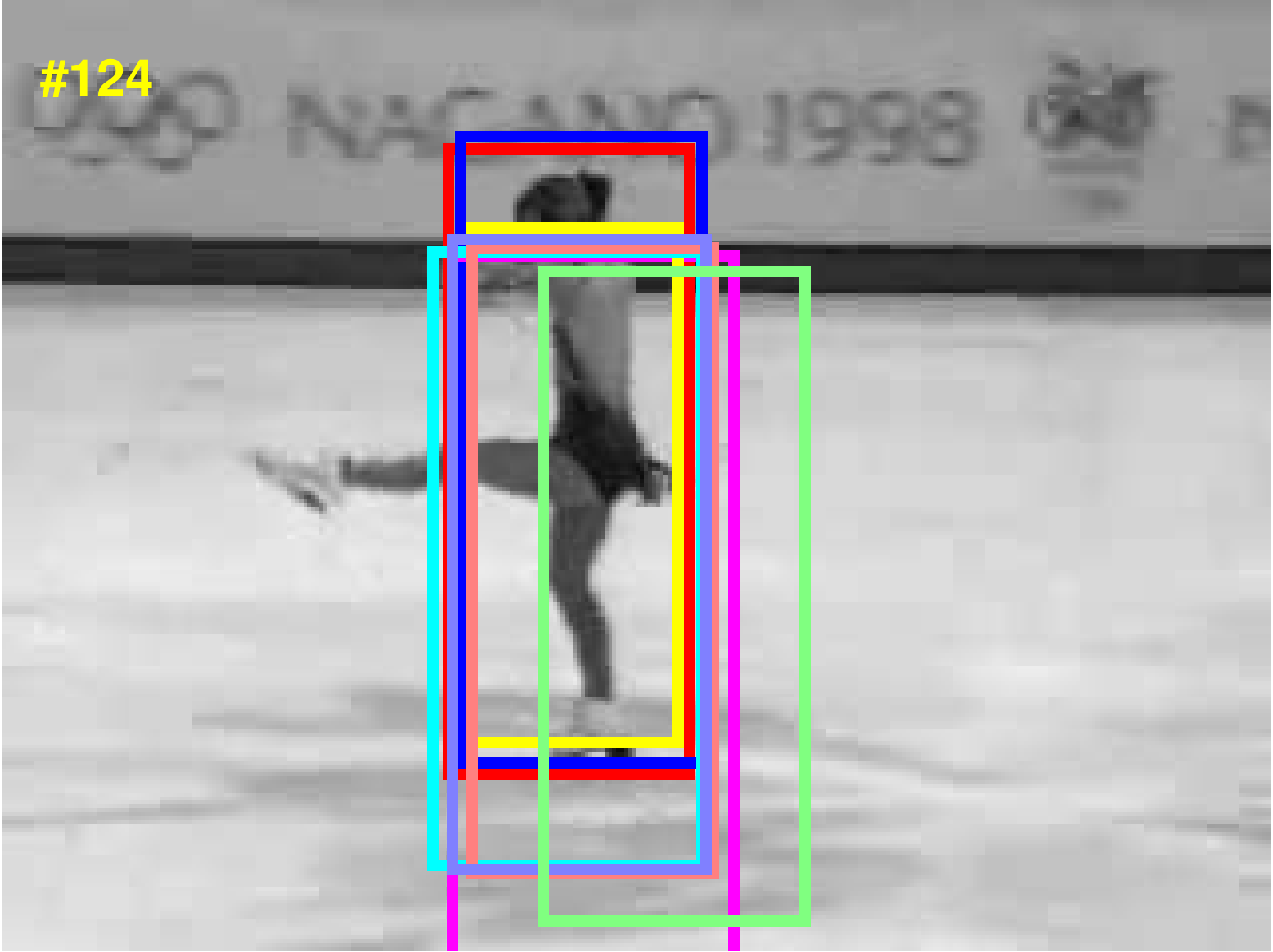}
	 \includegraphics[width=0.161\textwidth, height=0.12\textwidth]{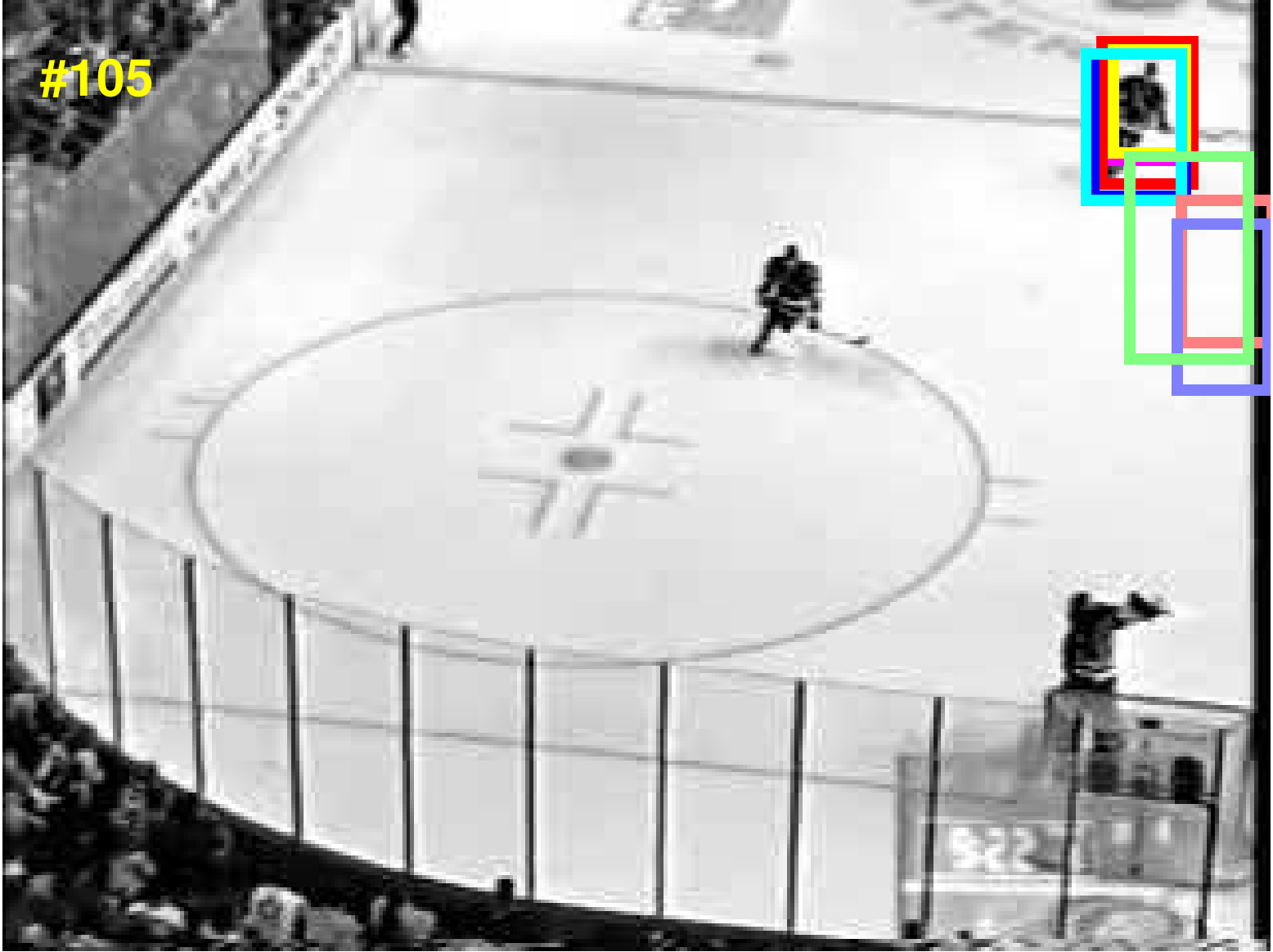}
	 \includegraphics[width=0.161\textwidth, height=0.12\textwidth]{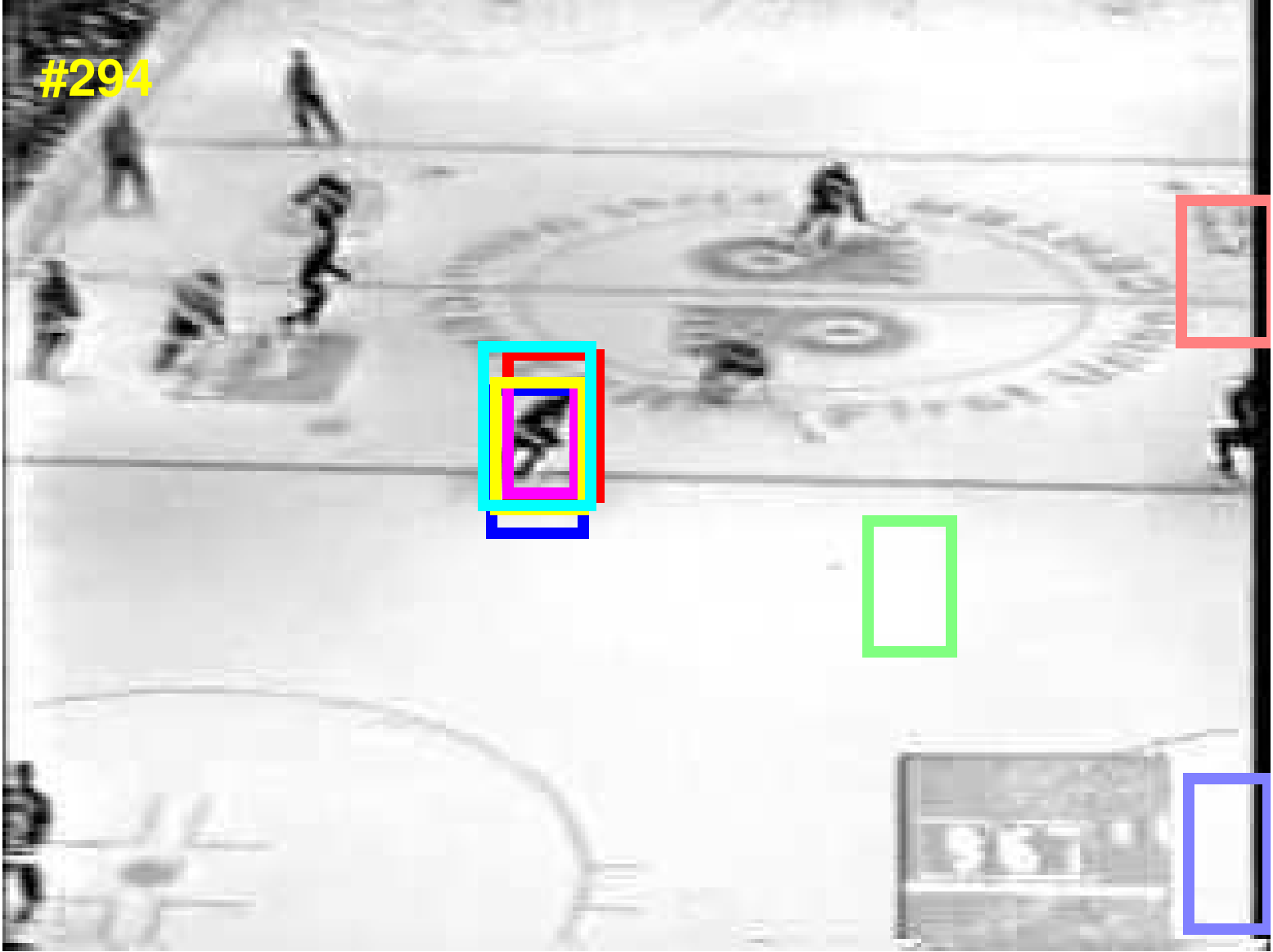}  \\

	 \includegraphics[width=0.161\textwidth, height=0.12\textwidth]{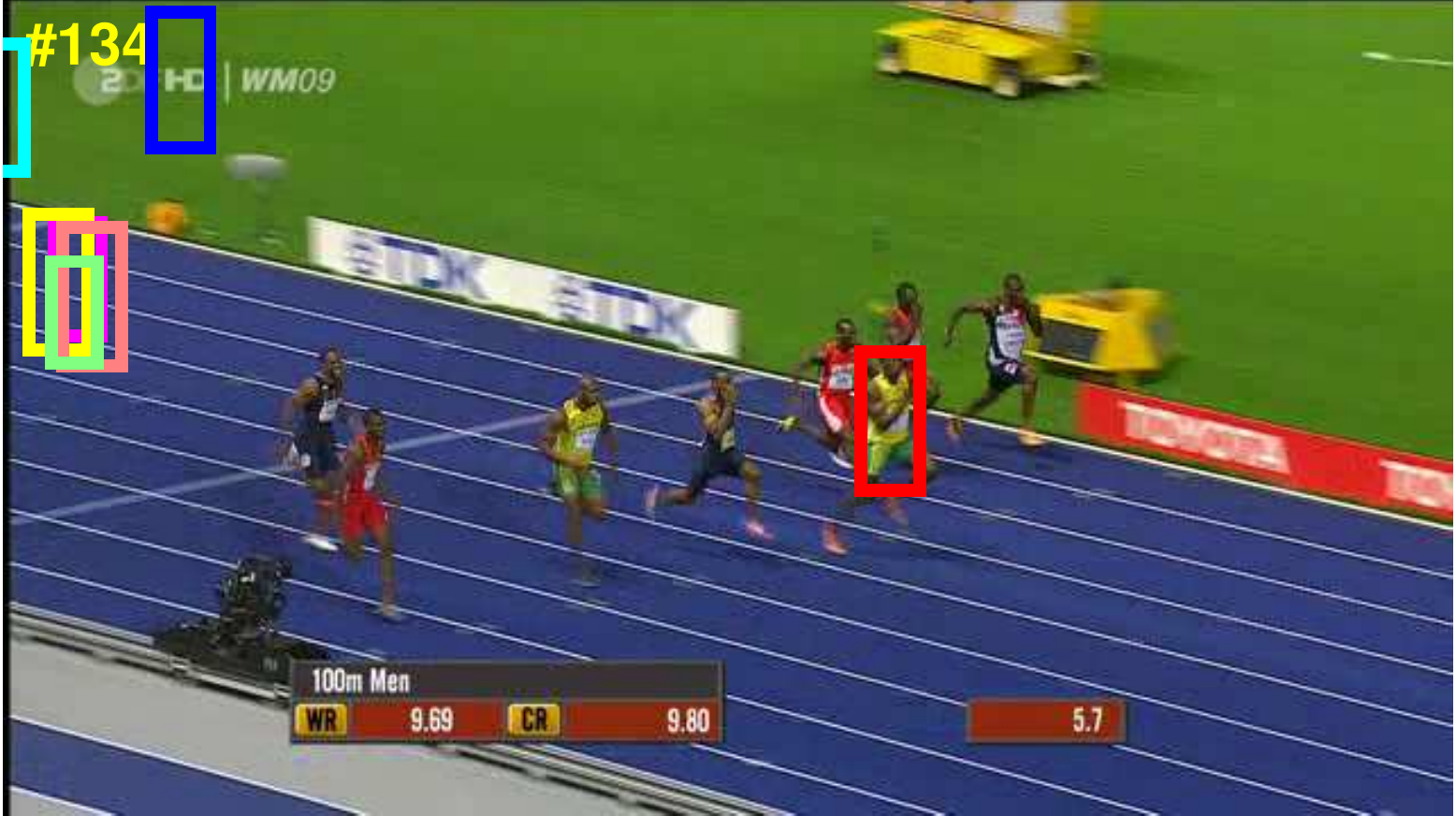}
	 \includegraphics[width=0.161\textwidth, height=0.12\textwidth]{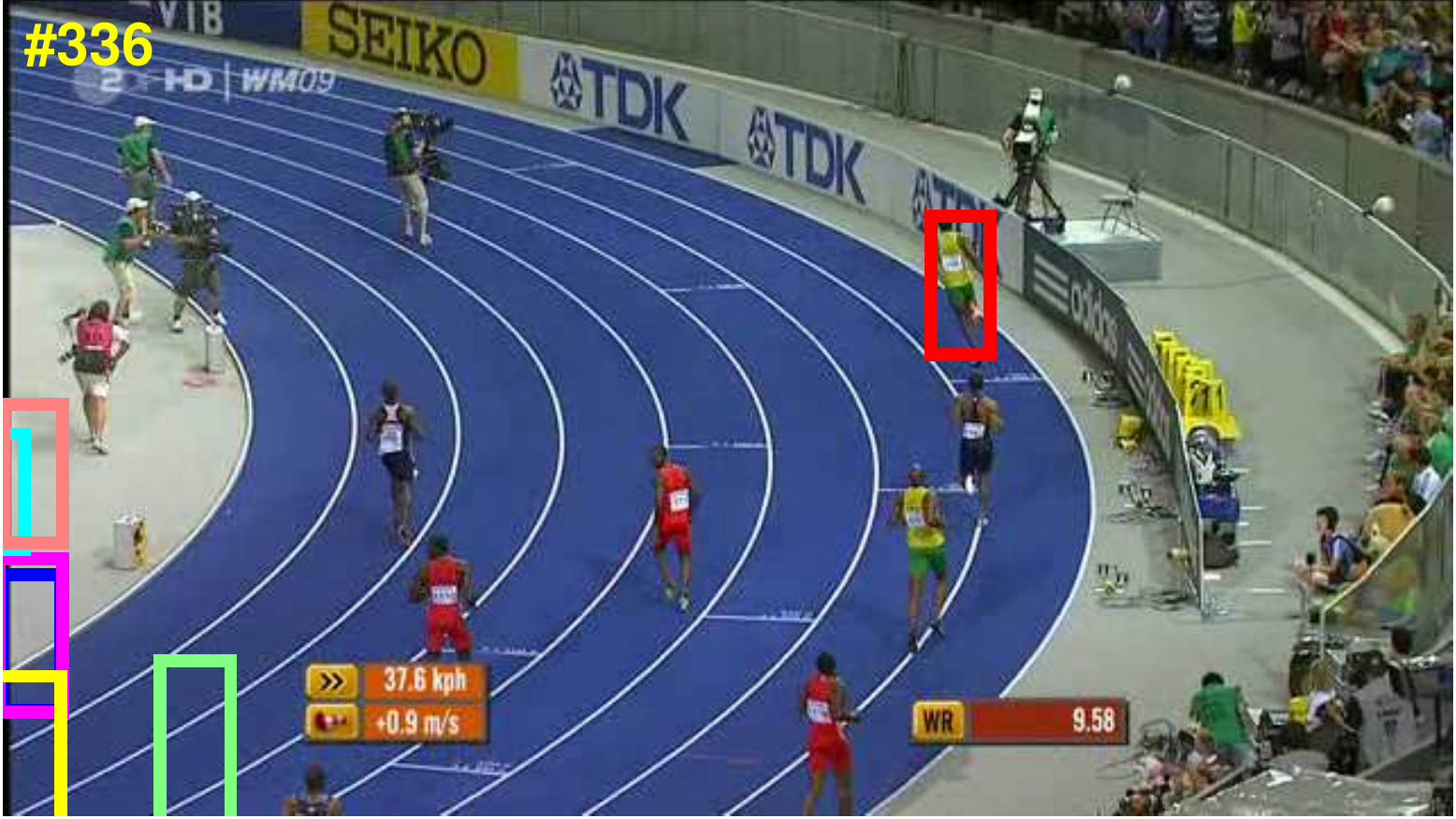}
	 \includegraphics[width=0.161\textwidth, height=0.12\textwidth]{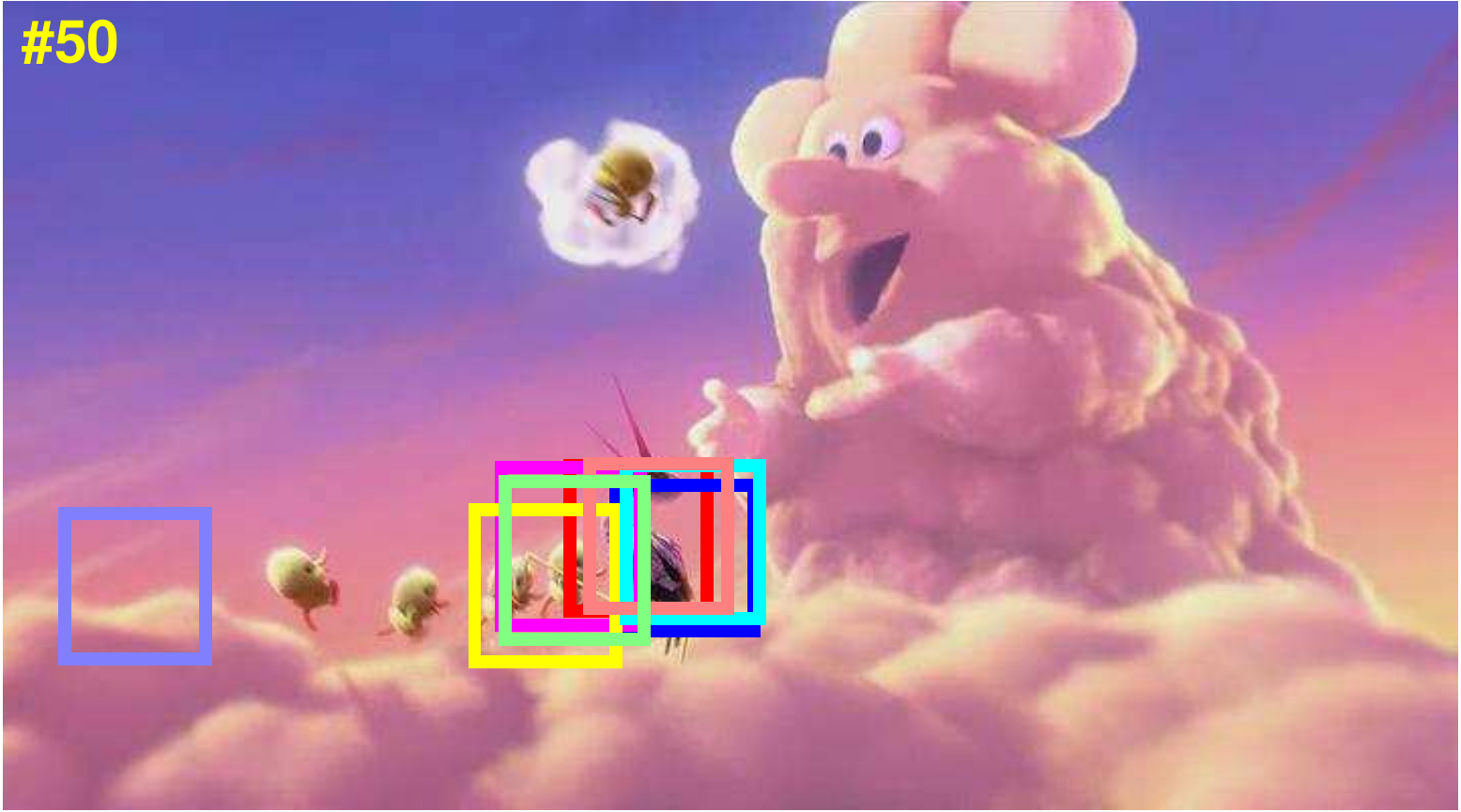}
	 \includegraphics[width=0.161\textwidth, height=0.12\textwidth]{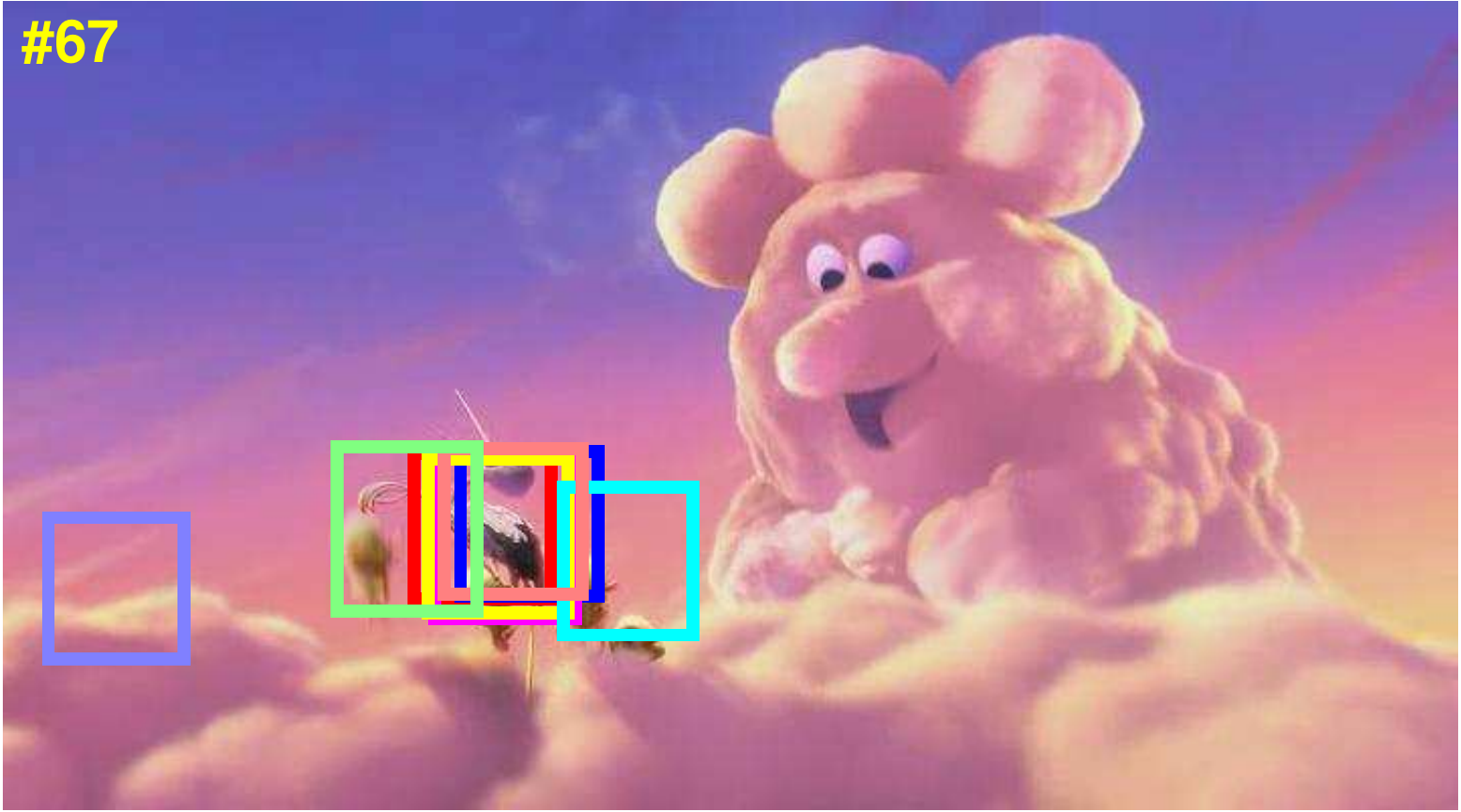}
	 \includegraphics[width=0.161\textwidth, height=0.12\textwidth]{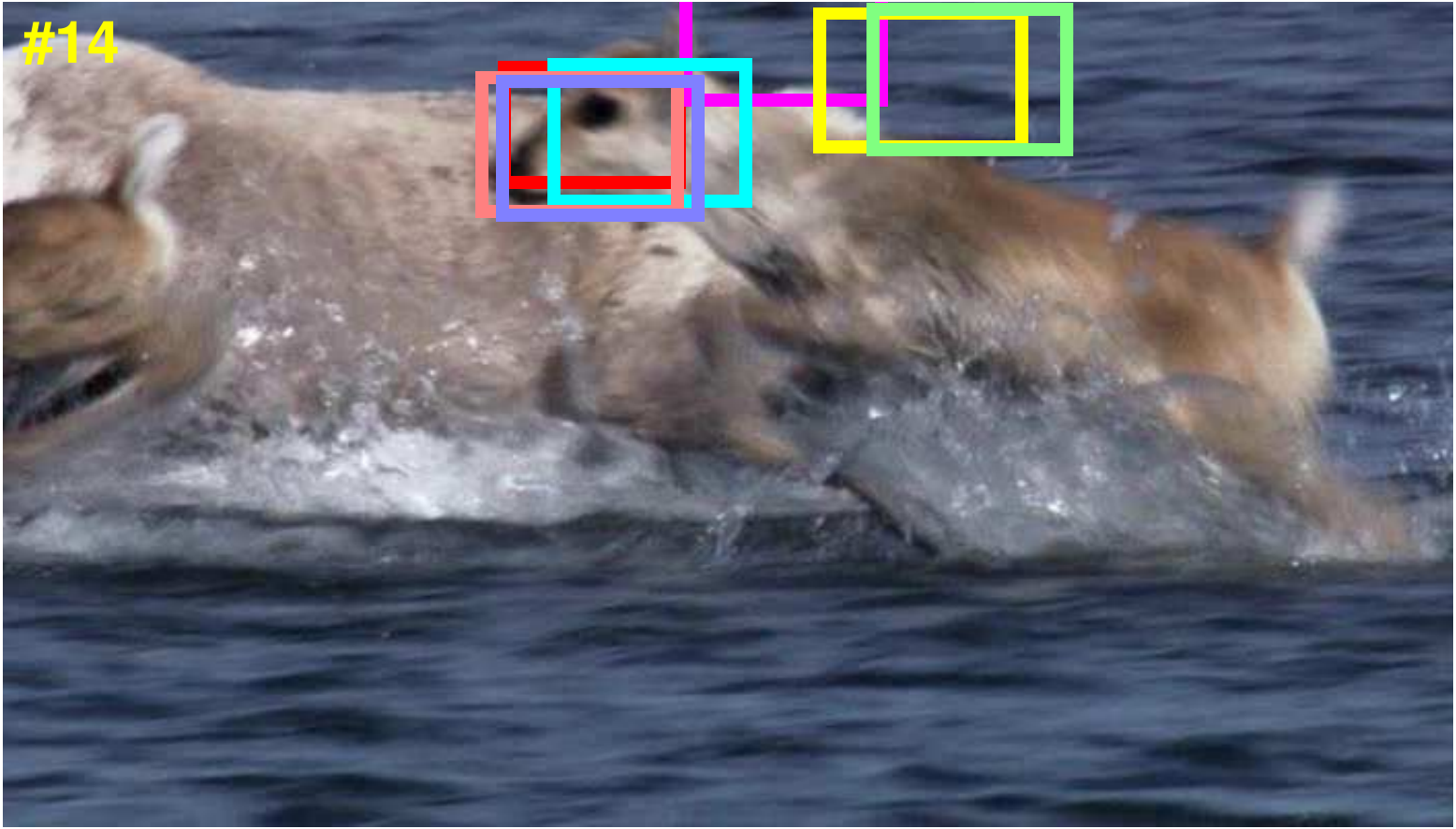}
	 \includegraphics[width=0.161\textwidth, height=0.12\textwidth]{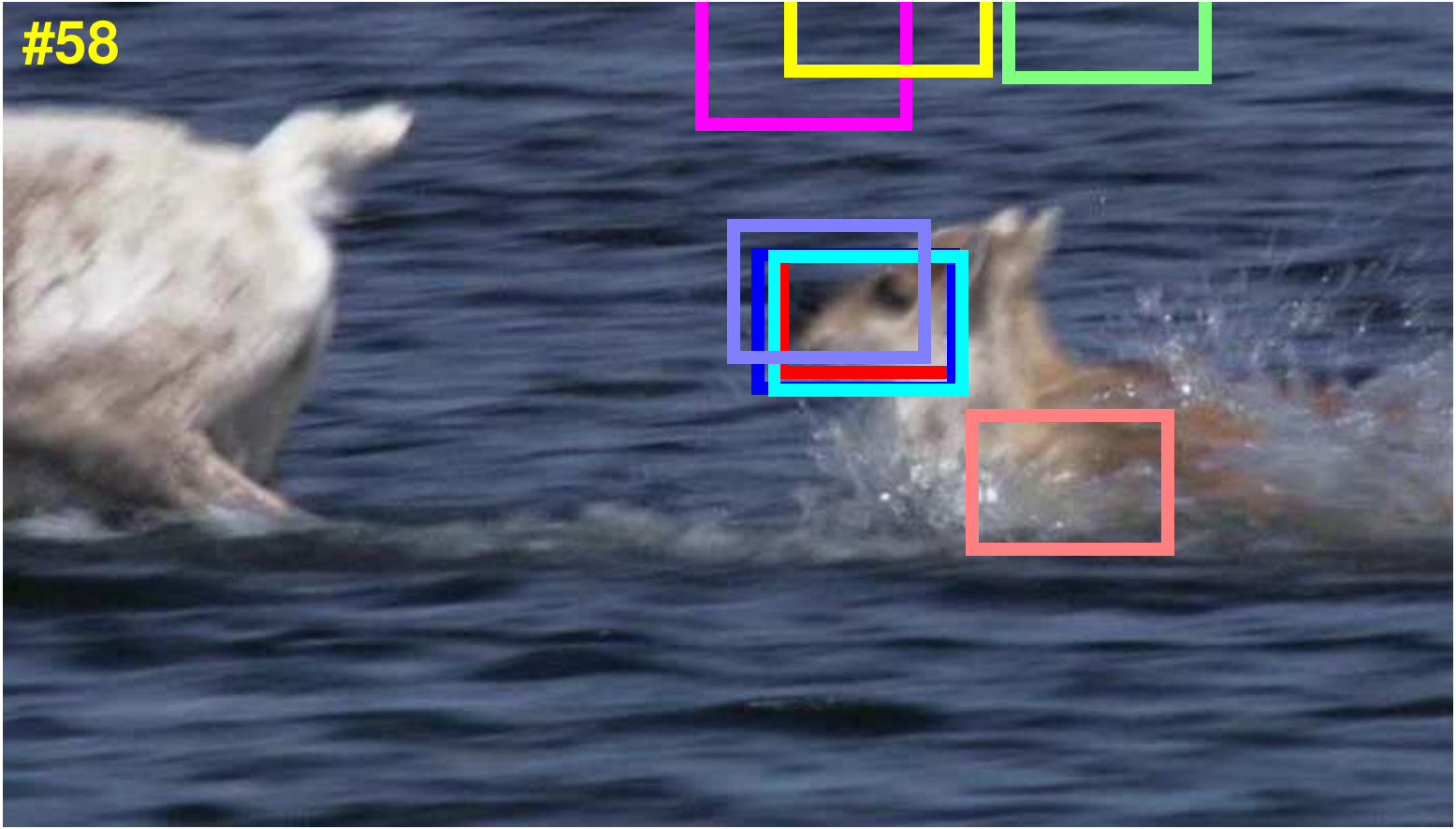}  \\

\vspace{.2cm}
  \includegraphics[width=7.5cm, height=.35cm]{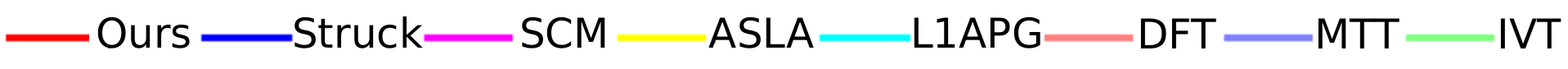}
\caption{Qualitative comparison on sequence david, girl, faceocc2, david3, woman, tiger2, dollar, box, board, shaking, fskater, iceball, bolt, bird, deer. }  \label{fig:trackBox}
\end{figure*}

\section{Conclusion}
We have presented an online feature learning based tracker in this work. The proposed tracker follows the classical feature learning pipeline, which consists of dictionary learning, feature encoding and spatial pooling. The online dictionary learning method is applied to account for the appearance variations of the tracking target. We also evaluate the roles of several commonly used dictionary learning as well as encoding approaches in the proposed tracking framework, and achieve similar conclusions with previous studies on image classification. When combined with Struck, the learned features help improve the tracking accuracy compared to traditional hand-crafted features. Experimental results on various challenging videos demonstrate that the proposed tracker outperforms the state-of-the-art.
Future work may take into consideration incorporating motion models of the target and tracking multiple objects.

{%
\small
    \bibliographystyle{ieee-cs}
    \bibliography{tracking}
}

\end{document}